\documentclass[10pt,twocolumn,letterpaper]{article}

\usepackage{iccv}
\usepackage{times}
\usepackage{epsfig}
\usepackage{graphicx}
\usepackage{amsmath}
\usepackage{amssymb}
\usepackage{mathtools}
\usepackage{bm}
\usepackage{booktabs}
\usepackage{lipsum}
\usepackage{multirow}
\usepackage{tabularx}
\usepackage{makecell}
\usepackage{blindtext}
\usepackage{url}
\usepackage{enumitem}
\setlist[itemize]{topsep={0pt},partopsep={0pt}}

\usepackage{caption}
\usepackage{xcolor}
\usepackage{colortbl}

\definecolor{citecolor}{HTML}{1976D2}
\definecolor{linkcolor}{rgb}{0.956,0.298,0.235} 
\definecolor{fontcolor}{rgb}{0.267,0.420,0.809} 

\definecolor{gray}{gray}{0.95}
\definecolor{cyan}{rgb}{0.831,0.901,0.945}

\usepackage[pagebackref,breaklinks,colorlinks,linkcolor=linkcolor,citecolor=citecolor,bookmarks=false]{hyperref}

\iccvfinalcopy 

\def\iccvPaperID{****} 

\ificcvfinal\fi

\begin{document}

\title{\mbox{Make-It-3D: High-Fidelity 3D Creation from A Single Image with Diffusion Prior}}


\author{
Junshu Tang$^1$${^{\dagger}}$ 
\quad Tengfei Wang$^2$${^{\dagger}}$ 
\quad Bo Zhang$^3$\footnotemark[3]
\quad Ting Zhang$^3$\\
\quad Ran Yi$^1$ 
\quad Lizhuang Ma$^1$\footnotemark[3]
\quad Dong Chen$^3$\\
{$^1$Shanghai Jiao Tong University \quad $^2$HKUST\quad $^3$Microsoft Research} \\
\url{https://make-it-3d.github.io/}
}


\twocolumn[{
\renewcommand\twocolumn[1][]{#1}
\maketitle
\centering
\vspace{-0.7cm}
 \includegraphics[width=\linewidth]{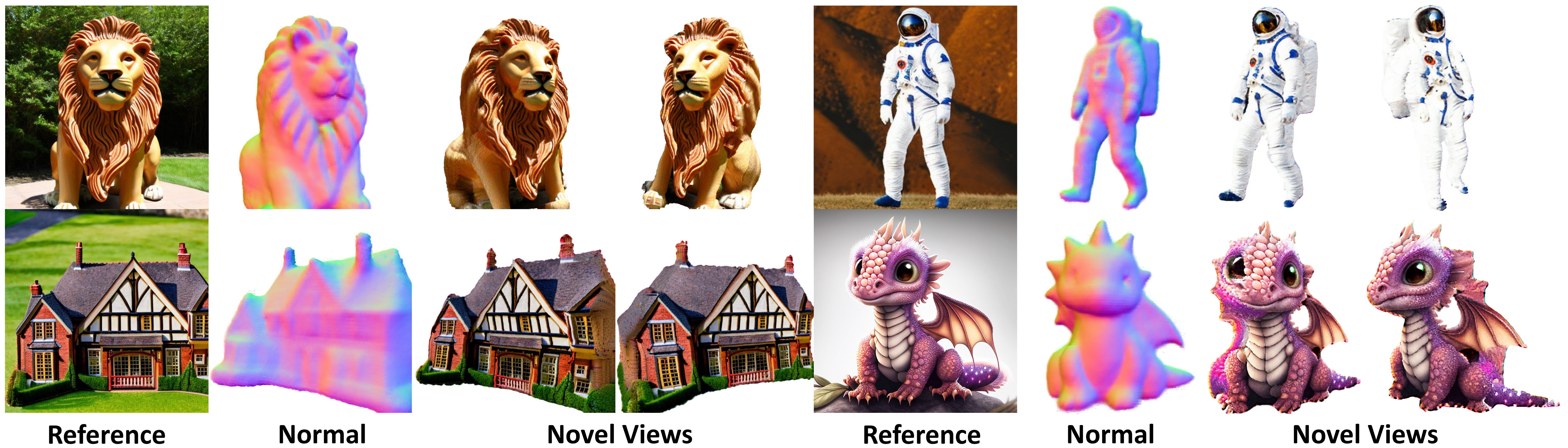}
 
 \captionsetup{type=figure}
\caption{\emph{Make-It-3D} can create high-fidelity 3D content from only a single image. We show the normal map and novel-view renderings of created 3D content, showcasing  fine geometry and faithful textures with stunning quality at novel views.}
\vspace{0.2cm}
\label{fig:teaser}
}]

{
  \renewcommand{\thefootnote}%
    {\fnsymbol{footnote}}
  \footnotetext[2]{Work is done during the internship at Microsoft Research.}
  \footnotetext[3]{Corresponding authors.}
}

\begin{abstract}
\vspace{-3.8mm}
In this work, we investigate the problem of creating high-fidelity 3D content from only a single image. This is inherently challenging: it essentially involves estimating the underlying 3D geometry while simultaneously hallucinating unseen textures. To address this challenge, we leverage prior knowledge from a well-trained 2D diffusion model to act as 3D-aware supervision for 3D creation. Our approach, \textbf{\emph{Make-It-3D}}, employs a two-stage optimization pipeline: the first stage optimizes a neural radiance field by incorporating constraints from the reference image at the frontal view and diffusion prior at novel views; the second stage transforms the coarse model into textured point clouds and further elevates the realism with diffusion prior while leveraging the high-quality textures from the reference image. Extensive experiments demonstrate that our method outperforms prior works by a large margin, resulting in faithful reconstructions and impressive visual quality. Our method presents the first attempt to achieve high-quality 3D creation from a single image for general objects and enables various applications such as text-to-3D creation and texture editing. 
\end{abstract}

\section{Introduction}

Given a single image as in Figure~\ref{fig:teaser}, 
how would the object portrayed in the image look like from a different perspective? Humans possess an innate ability to effortlessly imagine 3D geometry and hallucinate the appearance of novel views with a glance at the picture based on their prior knowledge about the world.  In this work, we aim to achieve a similar goal:  creating  high-fidelity 3D content from a real or artificially generated single image.  This will open up new avenues for artistic expression and creativity, such as bringing 3D effects to the fantasy images created by the cutting-edge 2D generative models like Stable Diffusion~\cite{rombach2022high}. 
By offering a more accessible and automated way to create visually stunning 3D content, we hope to engage a broader audience with the world of 3D modeling with ease.

The creation of 3D objects from a single image presents a significant challenge due to the limited information that can be inferred from a single viewpoint.  
One categories of works aim to produce 3D photo effect~\cite{niklaus20193d,shih20203d,jampani2021slide,tucker2020single} in the manner of  image-based rendering or single-view 3D reconstruction with neural rendering~\cite{wiles2020synsin,xu2022sinnerf,Rockwell2021}. However, these methods often struggle with reconstructing fine geometry and fall short of rendering in large views. Another line of research~\cite{nichol2022point,wang2022rodin,yu2021pixelnerf,xie2022high} projects the input image into the latent space of the pretrained 3D-aware generative networks. Despite their impressive performance, existing 3D generative networks mainly model objects from a specific class and are therefore incapable of handling general 3D objects. In our case, we aim for general 3D creation from an arbitrary image, yet constructing a sufficiently large and diverse dataset for estimating the novel views or building a powerful 3D foundation model for general objects remains insurmountable.

Unlike the scarcity of 3D models,   images are much more readily available, and recent advancements in diffusion models have sparked a revolution in 2D image generation~\cite{ramesh2022hierarchical,saharia2022photorealistic,rombach2022high,wang2022pretraining,Balaji2022eDiffITD}. Interestingly, we observed that well-trained image diffusion models can generate images under various views, which implies that they have already incorporated  3D knowledge. This has motivated us to explore the possibility of cultivating prior knowledge in a 2D diffusion model to reconstruct 3D objects. With diffusion prior,   we propose \emph{Make-It-3D}, a two-stage 
3D content creation method that can generate a high-fidelity 3D object with superior quality from only one image. 

In the first stage, we leverage diffusion prior to optimize a neural radiance field (NeRF)~\cite{mildenhall2021nerf}  by applying score distillation sampling (SDS)~\cite{poole2022dreamfusion}, and constrain this optimization with reference-view supervision.  
Different from prior text-to-3D works~\cite{poole2022dreamfusion,lin2022magic3d,metzer2022latent},  we focus on image-based 3D creation so that we need to prioritize the faithfulness to the reference image. However, we observed that while 3D models generated with SDS match text prompts well, they often fail to align faithfully with reference images since textual descriptions do not capture all object details. To address this issue, we go beyond SDS by
simultaneously maximizing the image-level similarity between the reference and the novel view rendering denoised by a diffusion model. Also, as images inherently capture more geometry-related information than textual descriptions, we can thus incorporate the depth of the reference image as an extra geometry prior to alleviate the shape ambiguity of NeRF optimization.

While the first stage generates a coarse model with plausible geometry, its appearance often deviates from the quality of the reference, exhibiting over-smooth textures and saturated colors~\cite{poole2022dreamfusion}. This has limited its overall realism, and it is imperative to further bridge the gap between  coarse model and  reference image. As texture is more critical than geometry for human perception in the context of high-quality rendering, we  choose to prioritize texture enhancement   in the second stage, while inheriting the geometry from the first stage. We refine the model by leveraging the availability of ground-truth textures for regions that are observable in the reference image. To achieve this, we export  the coarse NeRF model to textured point clouds   and project reference textures onto their corresponding areas in the point clouds. We then utilize diffusion prior to enhance the texture of the remaining points by jointly optimizing the point feature and a point cloud renderer, resulting in a clearly improved texture of the generated 3D model.

With  diffusion prior as multi-view supervision, our approach  can be applied to general objects  without being limited to specific categories. To evaluate the method, we create a benchmark consisting of 400 images including both real images and   generated images from 2D diffusion.  We evaluate the proposed method on public DTU dataset~\cite{aanaes2016large} and our benchmark, and extensive experiments show  a clear improvement over previous works. Furthermore, our method  enables a range of applications beyond image-to-3D creation such as texture editing  and high-quality text-to-3D creation.  
Our main contributions are summarized as:
\begin{itemize}[noitemsep,topsep=0pt]
    \item  We propose \emph{Make-It-3D}, a  framework to create   a high-fidelity 3D object  from a single image, using   a 2D diffusion model as 3D-aware prior. It does not require multi-view images for training and can be applied to any input image, whether it is real or generated.
    \item With a  two-stage creation scheme, \emph{Make-It-3D} represents the first work to achieve high-fidelity 3D creation for general objects. The resulting 3D models exhibit detailed geometry and realistic textures that accurately conform to the reference images.
    \item Beyond image-to-3D creation, our method enables multiple applications  such as high-quality text-to-3D creation and texture editing. 
\end{itemize}

\section{Related Work}
\noindent \textbf{Novel view synthesis from a few images.}
Early attempts~\cite{levoy1996light,gortler1996lumigraph,sun2018multi,park2017transformation} usually require dense observations of a scene from uniformly sampled poses. 
Recent emergence of implicit representations~\cite{sitzmann2019scene,mildenhall2021nerf}
significantly advances the synthesis quality of novel views, whereas they tend to find a degenerate solution when given only very few input views.
To enable novel view from   sparse  input views,
a growing body of works~\cite{lombardi2019neural,trevithick2021grf,niemeyer2022regnerf} hence turn to extra prior knowledge   as additional regularizations. 
PixelNeRF~\cite{yu2021pixelnerf} predicts a continuous neural  representation conditioned
on the input images rather than only leveraging input views for supervision.
DietNeRF~\cite{jain2021putting} penalizes a semantic consistency loss by minimizing distance between CLIP~\cite{radford2021learning} features of different views.
With recent rapid progress of diffusion models, 3DiM~\cite{watson2022novel} introduces a pose-conditional  diffusion model that generates a
novel view conditioned on a source view and a target pose.
RenderDiffusion~\cite{anciukevivcius2022renderdiffusion} presents a diffusion model for 3D generation that incorporates a triplane rendering mode into the denoiser.

\begin{figure*}[t]
    \centering
    \includegraphics[width=0.93\linewidth]{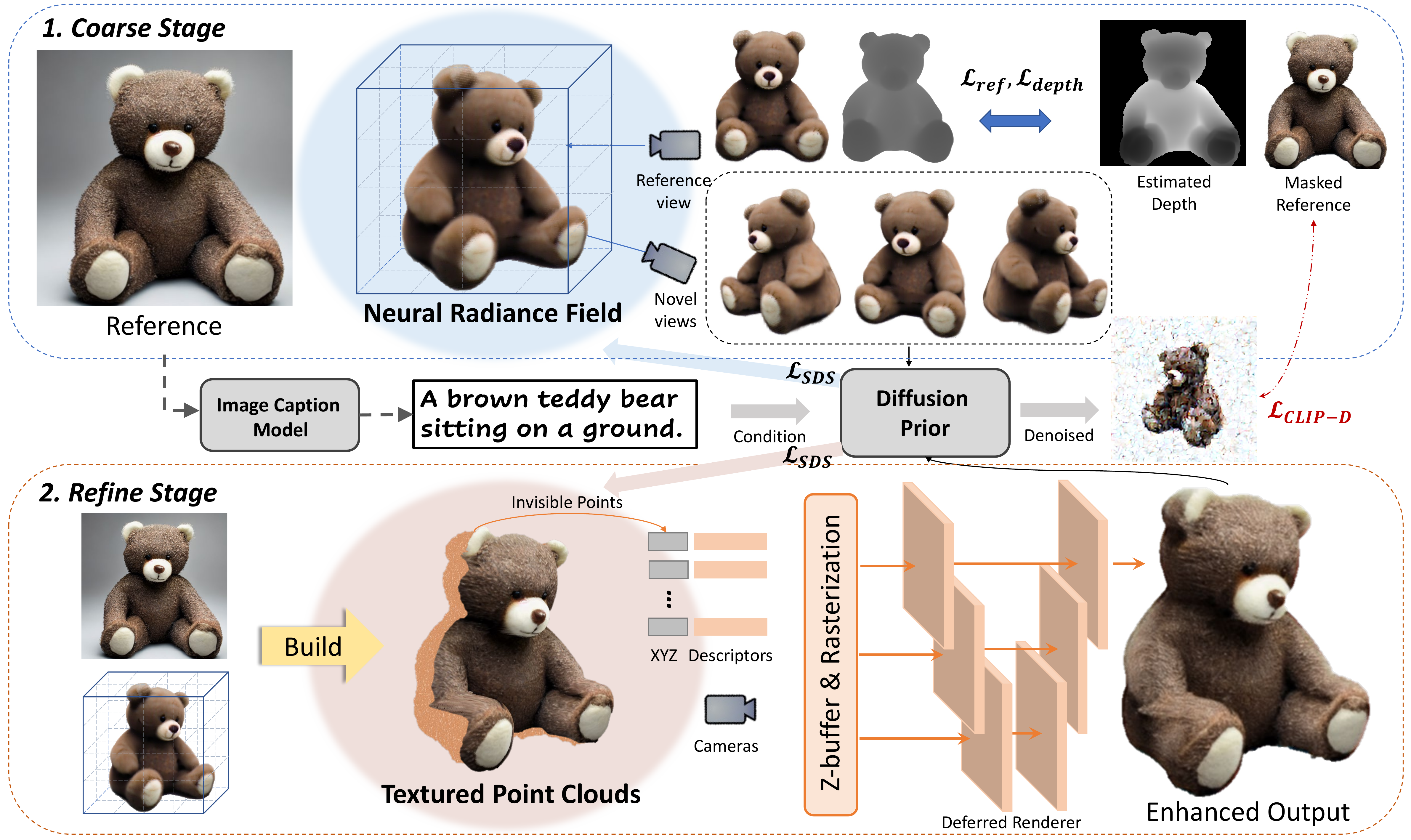}
    \caption{Overview architecture. We propose a two-stage  framework for creating  a high-quality 3D model from a reference image with diffusion prior (Sec.~\ref{sec:diffusion}). At the coarse stage, we optimize a NeRF for reconstructing the geometry of the reference image (Sec.~\ref{sec:coarse}). We further build textured point clouds from NeRF and the reference image, and jointly optimize the texture of invisible points and a learnable deferred renderer to generate realistic and view-consistent textures (Sec.~\ref{sec:refine}). }
    \label{fig:pipeline}
    \vspace{-3mm}
\end{figure*}

\noindent \textbf{Single-image 3D photography.}
Synthesizing novel views from a single image is quite challenging as it is a highly ill-posed problem, requiring precise geometry estimation and disocclusion of both geometry and texture. Increasing effort has been dedicated to this problem~\cite{zhou2016view,habtegebrial2018fast,liu2018geometry,olszewski2019transformable,park2017transformation,rematas2016novel, niklaus20193d,srinivasan2017learning}, many of which can only handle specific types.
Among them, a number of  methods rely on layered representations such as layered depth images~\cite{shade1998layered,tulsiani2018layer,shih20203d} and multi-plane images (MPIs)~\cite{srinivasan2019pushing,tucker2020single,li2020synthesizing,han2022single}.
For example,~\cite{tucker2020single}  predicts MPIs for view synthesis from single image without requiring ground
truth 3D. ~\cite{shih20203d} generates a 3D photo from a given RGB-D input through layered depth image  with inpainted color and depth. Yet such a  solution is limited by the number of planes and   sensitive to discontinuities. Subsequent efforts generalize MPIs to continuous 3D representations such as NeRF~\cite{li2021mine} and latent 3D point cloud~\cite{wiles2020synsin}.

\noindent \textbf{Lift 2D pretrained model to 3D.}
With the emergence of 
recent advances in modeling natural image manifold, how to exploit such powerful 2D pretrained model to recover 3D object structure
has received considerable research interest.
\cite{pan20202d}  attempts to reconstruct the 3D shape using pretrained 2D GANs.
Subsequently, some works~\cite{jain2022zero,lee2022understanding,mohammad2022clip}   explore zero-shot text-guided 3D content creation utilizing the guidance from CLIP~\cite{radford2021learning}.
Recent efforts such as DreamFusion~\cite{poole2022dreamfusion}, Magic3D~\cite{lin2022magic3d} and Score Jacobian Chaining~\cite{wang2022score} explore text-to-3D generation by exploiting a score distillation sampling (SDS) loss derived from a 2D text-to-image diffusion model instead, showing impressive results.
LatentNeRF~\cite{metzer2022latent}
proposes to use a shape prior to guide and assist the 3D generation directly in the latent space
of the diffusion model.
Prior works NeuralLift-360~\cite{xu2022neurallift} and NeRDi~\cite{deng2022nerdi} also leverage the generative prior for 3D reconstruction from a single view. 
Yet the reconstructed 3D model has limited quality and is poorly aligned with the input image.
In contrast, we propose a two-stage 3D synthesis framework with a relaxed SDS loss, yielding high-quality 3D representation faithful to the given input image.

\section{Method}
Generating novel views for general scenes or objects from only a single image is inherently challenging due to the difficulty of inferring both geometry and missing texture.
We therefore tackle this challenge by cultivating the dark knowledge of pretrained 2D diffusion models. Specifically, given an input image $\bm{x}$, we first hallucinate its underlying 3D representation, neural radiance field (NeRF), whose rendering appears as a plausible sample to a pretrained denoising diffusion model, and we constrain this optimization process with the texture and  depth supervision at the reference view. 
To further improve the rendering realism, we keep the learned geometry and enhance the textures with the reference image. As such, in the second stage, we 
lift the input image to 
textured point clouds and focus on refining the color of the points occluded in the reference view. 
We leverage prior knowledge of the text-to-image generative model and the text-image contrastive model for both stages.
In this way, we achieve a  faithful 3D representation of the input image with restored high-fidelity  texture and geometry. The proposed two-stage 3D learning framework is illustrated in Figure~\ref{fig:pipeline}. We will subsequently brief the preliminaries and then detail our method.

\subsection{Preliminaries}
\label{sec:diffusion}
Recent findings show that pretrained 2D generative models offer rich 3D geometry knowledge for their 2D generation samples. Notably, DreamFusion~\cite{poole2022dreamfusion} uses a text-to-image diffusion model to guide the optimization of the 3D representation. Let $\mathcal{G}_{\theta}(\beta)$ be the rendered image at the given viewpoint $\beta$, where $\mathcal{G}$ is the differentiable rendering function for the 3D representation parameterized by $\theta$ and is amenable to choice. DreamFusion optimizes the neural radiance field such that its multi-view renderings look like high-quality samples from a frozen diffusion model. 

Specifically, a diffusion model $\bm{\epsilon}_{\phi}$ introduces a random amount of noise $\bm{\epsilon}$ to the rendered image $\bm{x}_0:=\mathcal{G}_{\theta}(\beta)$ at different timestep $t$, \ie,
  $\bm{x}_t = \alpha_t \bm{x}_0 + \sigma_t \bm{\epsilon}$, where $\epsilon\sim \mathcal{N}(\bm{0},\bm{I})$; $\alpha_t$ and $\sigma_t$ define a noise schedule whose log signal-to-noise ratio $\lambda_t = \log[\alpha_t^2/\sigma_t^2] $ linearly decreases with the timestep $t$. A pretrained text-conditioned diffusion model is trained to reverse this noising process given the text embedding $\bm{y}$. To optimize the 3D representation parameters to render images as close as good generation samples, a \emph{score distillation sampling} (SDS) loss $\mathcal{L}_{\textnormal{SDS}}$ is introduced 
  to push rendered images toward higher density region conditioned on the text embedding. Specifically, $\mathcal{L}_{\textnormal{SDS}}$ computes the difference of predicted noise and the added noise as per-pixel gradient which is used to update the scene parameters, \ie,
\begin{equation}
    \begin{aligned}
    \nabla_{\theta}\mathcal{L}_{\textnormal{SDS}}(\phi, \mathcal{G}_{\theta}) = \mathbb{E}_{t, \bm{\epsilon}}\left[w(t)(\bm{\epsilon}_{\phi}(\bm{x}_{t};\bm{y},t)-\bm{\epsilon})\frac{\partial \bm{x}}{\partial\theta}\right],
    \end{aligned}
    \label{eq:sds}
\end{equation}
where $w(t)$ is a weight function of different noise levels.
It can be proved that this loss essentially measures the similarity between the image and the text prompt.
The diffusion model acts as a critic and the gradient of $\mathcal{L}_{\textnormal{SDS}}$ will not be back-propagated through the diffusion network, resulting in efficient computation. As training proceeds, the NeRF parameters are updated during which the 3D object gradually reveals its texture and geometry. In practice, it is found that using a diffusion model with a strong classifier-free guidance strength leads to higher-quality 3D samples.

While DreamFusion uses the Imagen~\cite{saharia2022photorealistic} to reverse the noising process at the pixel level, we use the publicly available Stable Diffusion~\cite{rombach2022high} that models the latent space of the VQ-VAE~\cite{van2017neural} with an encoder $\mathcal{E}$ and a decoder $\mathcal{D}$. Hence, the used diffusion model digests the latent $\bm{z}_0:=\mathcal{E}(\mathcal{G}_{\theta}(\beta))$ and the reconstructed latent $\hat{\bm{z}_0}$ can be mapped to image space through $\hat{\bm{x}}=\mathcal{D}(\hat{\bm{z}_0})$.

\subsection{Coarse Stage: Single-view 3D Reconstruction}
\label{sec:coarse}
As the first stage, we  reconstruct a coarse NeRF from the single reference image $\bm{x}$ with the diffusion prior constraining the novel views. Our optimization is expected to meet the following requirements simultaneously: 1) the optimized 3D representation should  closely resemble the rendering appearance of the input observation $\bm{x}$ at the reference view; 2) the novel view renderings should demonstrate consistent semantics with the input and appear  as plausible as possible; 3) the generated 3D model should exhibit  compelling geometry. In view of these, we randomly sample  the camera poses around the reference view and enforce constraints upon the rendered images $\mathcal{G}_{\theta}$ for both the reference view and unseen views.

\noindent\textbf{Reference view per-pixel loss.}
To encourage consistent appearance with the input image, we penalize the pixel-wise difference between the rendering and the input image at the reference view $\beta_{\textnormal{ref}}$:
\begin{equation}
    \begin{aligned}
        \mathcal{L}_{\textnormal{ref}} = \lVert \bm{x}\odot \bm{m} - \mathcal{G}_{\theta}(\beta_{\textnormal{ref}})\rVert_{1}.
    \end{aligned}
\end{equation}
Here we apply the foreground matting mask $\bm{m}$ to segment out the foreground as we empirically find that this eases the geometry reconstruction, which conforms to~\cite{yariv2020multiview}.

\begin{figure}
    \centering
    \includegraphics[width=\linewidth]{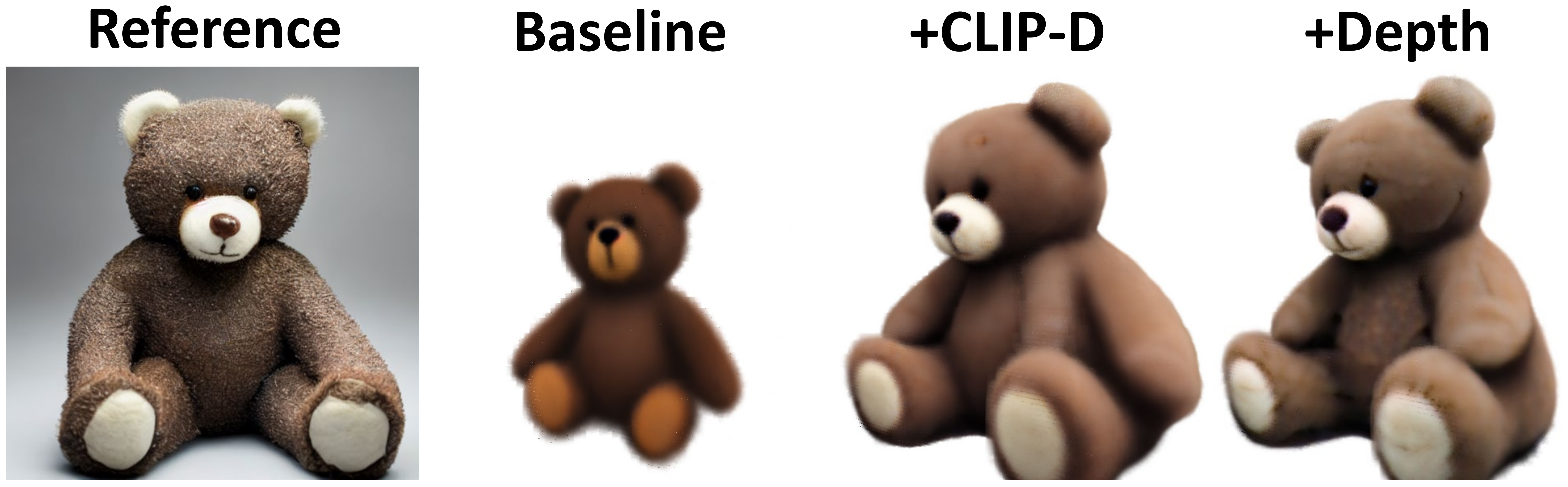}
    \caption{Analysis on the coarse stage. Baseline is a naive solution using only $\mathcal{L}_{\textnormal{SDS}}$ and $\mathcal{L}_\textnormal{ref}$, and it does not match the reference well. With $\mathcal{L}_{\textnormal{CLIP-D}}$, the result aligns better with the reference. The depth prior further improves faithfulness.}
    \label{fig:ab}
\end{figure}

\noindent\textbf{Diffusion prior.}
 Optimizing with the aforementioned losses can be unstable and may lead to implausible results, due to  the ill-posed nature of the problem. In order to encourage semantically plausible results, additional constraints are needed on the novel view rendering. To tackle this challenge, we resort to the diffusion prior. Prior works on text-to-3D~\cite{poole2022dreamfusion,lin2022magic3d} applied $\mathcal{L}_{\textnormal{SDS}}$ to leverage  text-conditioned diffusion models as 3D-aware prior. To utilize $\mathcal{L}_{\textnormal{SDS}}$ in our case,  we use an image captioning model~\cite{li2023blip}, to generate a detailed text description $y$ for the reference image. With the text prompt $y$, we can perform the SDS on the latent space of Stable Diffusion,
\begin{equation}
    \begin{aligned}
    \nabla_{\theta}\mathcal{L}_{\textnormal{SDS}}(\phi, \mathcal{G}_{\theta}) = \mathbb{E}_{t, \bm{\epsilon}}\left[w(t)(\bm{\epsilon}_{\phi}(\bm{z}_{t};\bm{y},t)-\bm{\epsilon})\frac{\partial \bm{z}}{\partial \bm{x}}\frac{\partial \bm{x}}{\partial\theta}\right],
    \end{aligned}
\end{equation}
where the noisy latent $\bm{z}_{t}$ is obtained form a novel view rendering $\bm{x}$ by Stable Diffusion encoder.

 However, as discussed 
 before, 
 $\mathcal{L}_{\textnormal{SDS}}$ essentially measures the similarity between the image and the given text prompt. While   $\mathcal{L}_{\textnormal{SDS}}$  can generate 3D models that are faithful to the text prompt, they do not align perfectly with the reference image (see baseline in Figure~\ref{fig:ab}), since text prompts cannot capture all object details. 
We go beyond this by a diffusion CLIP loss, denoted as $\mathcal{L}_{\textnormal{CLIP-D}}$, that additionally enforces the generated model to  match the reference image:
\begin{equation}
    \begin{aligned}
            \mathcal{L}_\textnormal{CLIP-D}(\mathcal{X}, \mathcal{G}_{\theta}(\beta)) = -\mathcal{E}_\textnormal{CLIP}(\mathcal{X}) \cdot \mathcal{E}_\textnormal{CLIP}(\mathcal{\hat{X}}_0(\beta, t)) ,
    \end{aligned}
\end{equation} 
where $\mathcal{E}_\textnormal{CLIP}(\cdot)$ is a  CLIP image encoder~\cite{radford2021learning}. Rather than directly measuring  CLIP loss on the rendered images $\mathcal{G}_{\theta}(\beta)$, we encode the novel view rendering $\mathcal{G}_{\theta}(\beta)$  to noisy latent  $\bm{z}_t$ and then denoise it to a clean image $\mathcal{\hat{X}}_0(\beta, t)$ with 2D diffusion. By imposing the similarity loss on denoised images sampled from diffusion models, we encourage the rendering   to align  with the reference image, while resembling high-quality samples from a frozen diffusion.
 
In detail, we do not optimize  $\mathcal{L}_\textnormal{CLIP-D}$ and $\mathcal{L}_\textnormal{SDS}$ at the same time. We use $\mathcal{L}_\textnormal{CLIP-D}$ at
small  timesteps and switch to $\mathcal{L}_\textnormal{SDS}$ at large timesteps. More details and analysis are  in the \textit{Supplement}. 
Combining $\mathcal{L}_\textnormal{SDS}$ and $\mathcal{L}_{\textnormal{CLIP-D}}$,  
our diffusion prior  ensures that the resulting 3D model appears  visually appealing and plausible while also conforming to the given image (see Figure~\ref{fig:ab}).


\begin{figure}
    \centering
    \includegraphics[width=\linewidth]{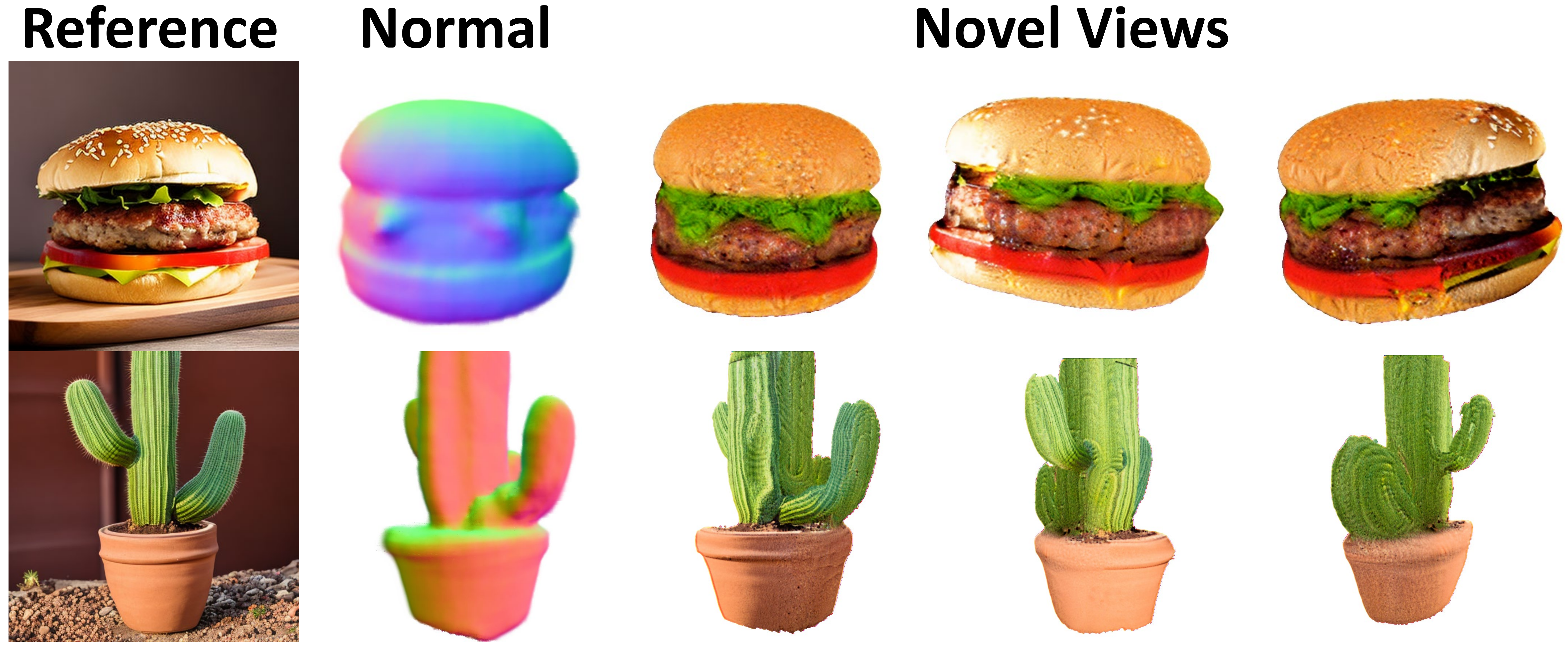}
    \caption{$360^\circ$ object reconstruction from real images. }
    \label{fig:360}
\end{figure}

\noindent\textbf{Depth prior.} Nonetheless, even if the rendered image appears meaningful to the diffusion model, there still exists shape ambiguity that brings about issues such as sunken faces, over-flat geometry~\cite{poole2022dreamfusion} or depth ambiguity (see Figure~\ref{fig:ab}). 
We mitigate these by leveraging depth prior learned from abundant external images and directly enforcing the supervision in 3D. To be specific, we utilize an off-the-shelf single-view depth estimator~\cite{Ranftl2021} to estimate the depth $d$ for the input image. While the estimated depth may not accurately characterize the geometric detail, it suffices to ensure plausible geometry and resolve most of the ambiguity. To account for the inaccuracy and the scale mismatch in $d$, akin to~\cite{deng2022nerdi}, we regularize the negative Pearson correlation between the estimated depth and the depth $d(\beta_{\textnormal{ref}})$ modeled by NeRF at the reference viewpoint, \ie,
\begin{equation}
    \begin{aligned}
        \mathcal{L}_{\textnormal{depth}}= -\frac{\textnormal{Cov}(d(\beta_{\textnormal{ref}}), d)}{\textnormal{Var}(d(\beta_\textnormal{ref}))\textnormal{Var}(d)},
    \end{aligned}
\end{equation}
where $\textnormal{Cov}(\cdot)$ denotes the covariance, $\textnormal{Var}(\cdot)$ computes the standard deviation.
With this regularization, the NeRF depth estimation is encouraged to be linearly correlated with the depth prior.

\noindent\textbf{Overall training.}
The overall loss can be formulated as a combination of $ \mathcal{L}_\textnormal{ref}$, $  \mathcal{L}_\textnormal{SDS}$, $\mathcal{L}_\textnormal{CLIP-D}$  and $\mathcal{L}_\textnormal{depth}$.To stabilize the optimization process, we adopt a progressive training strategy, where we start  with a narrow range of views near the reference view and gradually expand the range during training. With progressive training, we can achieve a  $360^\circ$ reconstruction of an object, as shown in Figure~\ref{fig:360}.
 

\begin{figure}
    \centering
    \includegraphics[width=\linewidth]{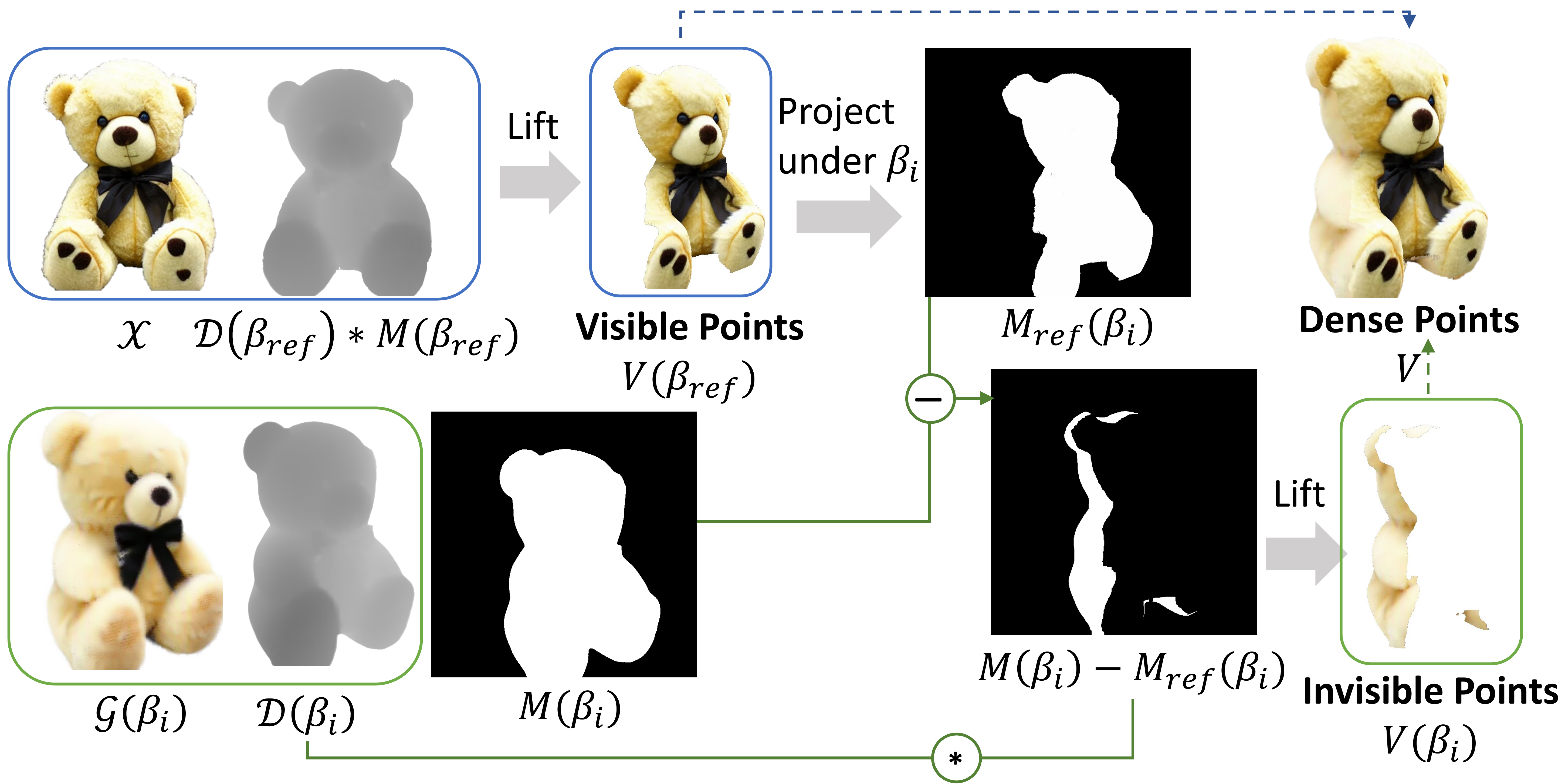}
    \caption{Illustration of textured point cloud building. We aim at building dense points and texturing visible points using reference image, and invisible points from NeRF.   } 
    \label{fig:mapping}
\end{figure}

\subsection{Refine Stage:  Neural Texture Enhancement}
\label{sec:refine}
\begin{figure}
    \centering
    \includegraphics[width=\linewidth]{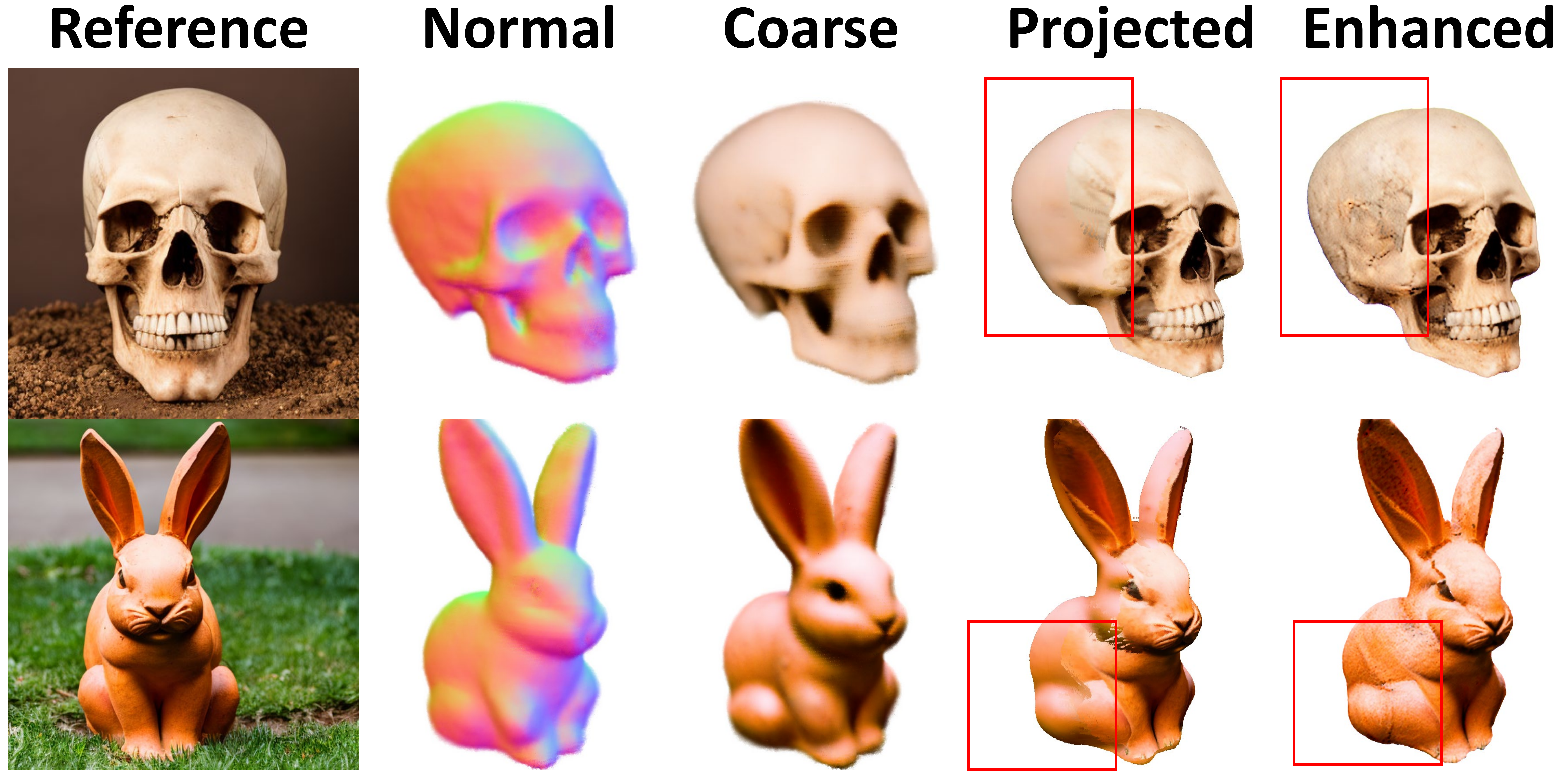}
    \caption{Visualization of neural texture enhancement. We project the reference textures to the coarse model and  enhance the remaining regions to high-frequency details consistent with the reference. Best viewed with zoom-in.}
    \label{fig:enhance}
    \vspace{-3mm}
\end{figure}

After the coarse stage, we obtained a 3D model with plausible geometry, but it often displays coarse textures   that can bottleneck  the overall quality  in Figure~\ref{fig:enhance}. Further refinement is thus desired for high-fidelity 3D models. Given that humans are more discerning when it comes to texture quality than geometry,   we prioritize texture enhancement while preserving the geometry of the coarse model. 

Our key insight for texture enhancement is that for a novel view, certain pixels can be observable in both the novel and reference views. Consequently, we can exploit this overlap to project the high-quality texture of the reference image onto the corresponding areas of the 3D representation. We then focus on enhancing the textures of regions that are occluded in the reference view. 

While NeRF is a suitable representation in the coarse stage as it can  handle topological changes continuously, projecting the reference image onto it is challenging. We thus opt to  export the neural radiance field to an explicit representation, specifically point clouds.   Compared to the noisy mesh exported by marching cube, point clouds offer a cleaner and more straightforward projection. 

\noindent\textbf{Textured point cloud building.}   A naive attempt to build point clouds is to render multi-view RGBD images from  NeRF and lift them to textured points in 3D space. However, we found this simple method leads to noisy point clouds due to the conflict among different views: a 3D point may possess different RGB colors in NeRF rendering under different views~\cite{xie2022high}.   We thus propose an iterative strategy to build  clean  point clouds from multi-view observations.

As in Figure~\ref{fig:mapping}, we first build point clouds from the reference view  $\beta_\textnormal{ref}$   according to the rendered 
  depth   $\mathcal{D}(\beta_\textnormal{ref})$ and  alpha mask $M(\beta_\textnormal{ref})$ of NeRF, 
\begin{equation}
    V(\beta_\textnormal{ref}) = R_\textnormal{ref}K^{-1}\mathcal{P}(\mathcal{D}(\beta_\textnormal{ref}) * M(\beta_\textnormal{ref}))),
\end{equation}
where $R_\textnormal{ref}$ and $K$ are the extrinsic and intrinsic matrices of the camera, and $\mathcal{P}$ denotes depth-to-point projection. These points are visible under the reference view and thus colorized with  ground-truth textures.  For the projection of the remaining views $\beta_i$, it is important to   avoid introducing  points that overlap with existing   points   but have conflicting colors.
To this end, we project the existing points $V(\beta_\textnormal{ref})$ to   the novel view $\beta_i$ to yield  a mask  indicating the presence of existing points. With this mask as guidance, we only lift those points $V(\beta_{i})$ that have not been observed yet, as shown in Figure~\ref{fig:mapping}.
These invisible points are then initialized with coarse textures from NeRF rendering $\mathcal{G}(\beta_i)$ and  integrated into the  dense point clouds.

\noindent\textbf{Deferred point cloud rendering.}
So far, we have built a set of textured point clouds  $V=\{ V(\beta_\textnormal{ref}), V(\beta_1), ..., V(\beta_{N})\}$. Though  $ V(\beta_\textnormal{ref})$   already have high-fidelity textures projected from the reference image,   the other points that are occluded in the reference view still suffer smooth textures from the coarse NeRF, as shown in Figure~\ref{fig:enhance}. To enhance the texture, we optimize the texture of the other points  and constrain novel-view rendering  with diffusion prior.  Specifically, we optimize a 19-dimensional descriptor $F$ for each point, whose first three dimensions are initialized with the initial RGB colors.
To avoid noisy colors and bleeding artifacts~\cite{aliev2020neural}, we  adopt a multi-scale deferred rendering scheme. In particular, given a novel view  $\beta$, we rasterize the point cloud  V for $K$ times to obtain $K$   feature maps $I_{i}$ with varying  sizes of $[W / 2^{i}, H / 2^{i}]$, where $i\in[0, K)$. These feature maps are then concatenated and rendered into an image $\textbf{I}$ using a U-Net renderer $\mathcal{R}_{\theta}$~\cite{aliev2020neural} that is jointly optimized:
\begin{equation}
    \begin{aligned}
        &I_{i}(\beta) = \mathcal{S}(i, V, F, \beta),~i\in[0, K), \\
        &\mathbf{I}(\beta) = \mathcal{R}_{\theta}(I_{0}(\beta), I_{1}(\beta), ..., I_{K-1}(\beta)),
    \end{aligned}
\end{equation}
where $\mathcal{S}$ is a differentiable point rasterizer. The objective of the texture enhancement process is similar to that of the geometry creation discussed in Sec.~\ref{sec:coarse}, but we additionally include a regularization term that penalizes large differences between the optimized texture and the initial texture. 

\begin{figure*}
    \centering
    \includegraphics[width=\linewidth]{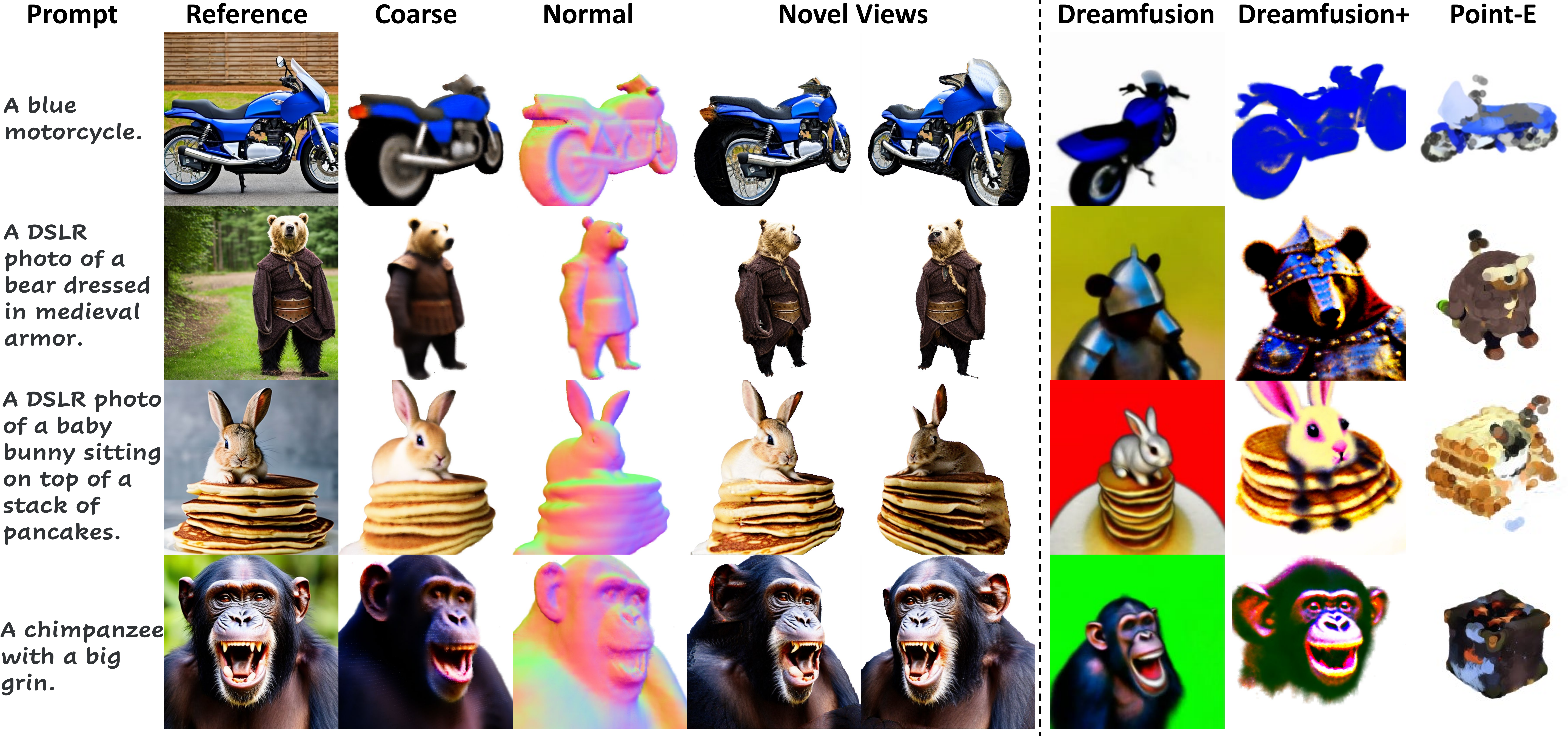}
    \caption{Qualitative comparison on the test benchmark with two diffusion-based 3D content creation models, Dreamfusion and Point-E. We show our results with high-fidelity geometry and texture. The results of Dreamfusion are from its website.}
    \label{fig:synthetic}
    \vspace{-3mm}
\end{figure*}

\begin{figure*}
    \centering
    \includegraphics[width=\linewidth]{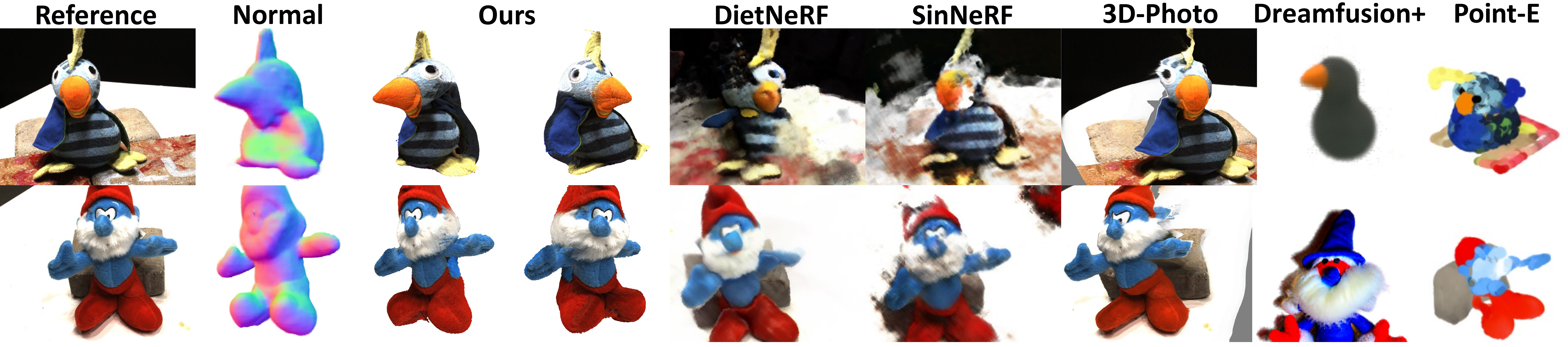}
    \caption{Qualitative comparison of novel view synthesis on DTU  with state of the arts.
    Our method generates sharper and more plausible details in both geometry and texture.}
    \label{fig:dtu}
    \vspace{-3mm}
\end{figure*}

\begin{table}[t]
\centering\footnotesize
\begin{tabular}{l|c|ccc}
\toprule
 &Views& LPIPS$\downarrow$  & Contextual$\downarrow$ & CLIP$\uparrow$ \\ \midrule
DietNeRF\protect~\cite{jain2021putting} & 3& 0.1831 & 5.34 & 64.90$\%$ \\
SinNeRF\protect~\cite{xu2022sinnerf} & 1& 0.2059 & 4.28 & 73.24$\%$ \\
DreamFusion+\protect~\cite{poole2022dreamfusion} & 1& 0.4075  & 2.15 &{82.81$\%$} \\
Point-E\protect~\cite{nichol2022point} & 1& -  & 2.23 & 71.31$\%$ \\
3D-Photo\protect~\cite{shih20203d} & 1& 0    & 3.43 & {87.65$\%$}\\
\midrule
Ours-coarse & 1& 0.1427& 1.74& 87.50$\%$ \\
Ours-enhanced & 1& \textbf{0.0908} & \textbf{1.59}& \textbf{95.65$\%$}\\
\bottomrule
\end{tabular}
\caption{Quantitative comparison on DTU. We compute LPIPS under the reference view, and  other two metrics under novel views.  LPIPS   of Point-E is not reported due to the lack of a defined reference view.}
\label{tab:dtu}
\end{table}

\begin{table}[t]
\centering\footnotesize
\begin{tabular}{l|ccc}
\toprule
 & LPIPS$\downarrow$  & Contextual$\downarrow$ & CLIP$\uparrow$ \\ \midrule
DreamFusion+\protect~\cite{poole2022dreamfusion}& 0.5649 & 3.07& 84.08$\%$ \\
Point-E\protect~\cite{nichol2022point} & - & 5.37& 64.36$\%$\\
\midrule
Ours-coarse& 0.2354 & 1.98 & 89.06$\%$  \\
Ours-enhanced& \textbf{0.0780} & \textbf{1.33} & \textbf{95.12$\%$} \\
\bottomrule
\end{tabular}
\caption{Quantitative comparison on the test benchmark.}
\label{tab:synthetic}
\vspace{-5mm}
\end{table}

\section{Experiments}
\subsection{Implementation Details}
\noindent\textbf{NeRF rendering.}
We use the multi-scale hash encoding from Instant-NGP~\cite{muller2022instant} to implement the NeRF representation in the coarse optimization stage, which enables neural rendering  at a computational cost. 
Similar to Instant-NGP, we maintain an occupancy grid to enable efficient ray sampling by skipping empty space. 
Additionally, we adopt several shading augmentations on the rendered images, such as Lambertian and normal shading, akin to~\cite{poole2022dreamfusion}.

\noindent\textbf{Point cloud rendering.} For deferred rendering, we use a 2D  U-Net architecture with gated convolutions~\cite{yu2019free}. 
The dimension of the point descriptor is 19, where the first 3 dimensions are initialized RGB colors and the remaining dimensions are randomly initialized. We also set a learnable descriptor for the background. 

\noindent\textbf{Camera setting.} Following the camera sampling method used in~\cite{poole2022dreamfusion}, we randomly sample novel views with a 75$\%$ probability and sample the pre-defined reference view with a 25$\%$ probability.  
We also randomly enlarge the FOV when rendering with NeRF, following~\cite{lin2022magic3d}. 


\noindent\textbf{Score distillation sampling.} We randomly sample $t$ from 200 to 600, and set $w(t)$ as a uniform weighting depending on the timestep. We also use classifier-free guidance with a guidance weight $\omega: \hat{\epsilon}_{\phi}(z_{t};y,t) = (1+\omega)\epsilon_{\phi}(z_t;y,t) - \omega\epsilon_{\phi}(z_{t};t)$. 
Our method aims to align the created 3D model with the input image, and we use a  guidance weight $\omega = 10$.

\noindent\textbf{Training speed.} We use Adam~\cite{kingma2014adam}  with a learning rate of 0.001 for both stages. The coarse stage is trained for 5,000 iterations at a rendering resolution of 100$\times$100. The refine stage then takes another 5,000 iterations at a rendering resolution of 800$\times$800. The entire training process takes approximately 2 hours on a single Tesla 32GB V100 GPU.



\noindent\textbf{Test Benchmark.} To the best of our knowledge, we are the first method  focusing  on high-fidelity 3D creation from an arbitrary  image. So we build a test benchmark consisting of 400 images, comprising  both real images and  images generated by Stable Diffusion~\cite{rombach2022high}.
Each image in the benchmark is accompanied by a foreground alpha mask, an estimated depth map, and a  text prompt.   The text prompts for real images are obtained from  an  image caption model~\cite{li2023blip}. We will make   this test benchmark publicly available.
 

\subsection{Comparisons with the State of the Arts}

\noindent\textbf{Baselines.}
We compare our method with five representative baselines. 1)  DietNeRF~\cite{jain2021putting},   a few-shot NeRF.   We train it with three input views. 2) SinNeRF~\cite{xu2022sinnerf},   a single-view NeRF method. 3) DreamFusion~\cite{poole2022dreamfusion}. As it is originally conditioned on text prompts, we also modify it with image reconstruction loss at the reference view, referred as   \emph{DreamFusion+}   for fair comparison. 4) Point-E~\cite{nichol2022point},   point cloud generation  conditioned on   image.   5) 3D-Photo~\cite{shih20203d}, depth-based image warping and inpainting method.

\noindent\textbf{Qualitative comparison.} 
We first compare our method with 3D generation baselines, where DreamFusion and DreamFusion+ leverage 2D diffusion as 3D prior and PointE is a 3D diffusion model.  As shown in Figure~\ref{fig:synthetic}, their generated models fail to align faithfully with the reference image and suffer smooth textures. In contrast, our method produces high-fidelity 3D models with  fine geometry and realistic textures.  Figure~\ref{fig:dtu} shows additional comparison on novel view synthesis.  SinNeRF and DietNeRF encounter difficulties in reconstructing complex objects due to the lack of multi-view supervision.   3D-Photo  fails to reconstruct underlying geometry and  produces visible artifacts in large views. In comparison, our method achieves remarkably faithful geometry and visually pleasing textures under novel views.

\noindent\textbf{Quantitative comparison.} 
A compelling generated 3D model should  closely resemble the input image at the reference view, and demonstrate consistent semantics with the reference under novel views.  We evaluate these two aspects using the following metrics: 1) LPIPS~\cite{zhang2018unreasonable}, which assesses the  
 reconstruction quality at the reference view, 2) contextual distance~\cite{mechrez2018contextual}, which measures pixel-level similarity between novel-view  rendering and the reference, and 3) CLIP score~\cite{radford2021learning}, which evaluates the semantic similarity between the novel view and the reference. As shown in Table~\ref{tab:dtu} and Table~\ref{tab:synthetic}, our approach substantially outperforms  baselines in terms of both reference-view and novel-view  quality.

\begin{figure}
    \centering
    \includegraphics[width=\linewidth]{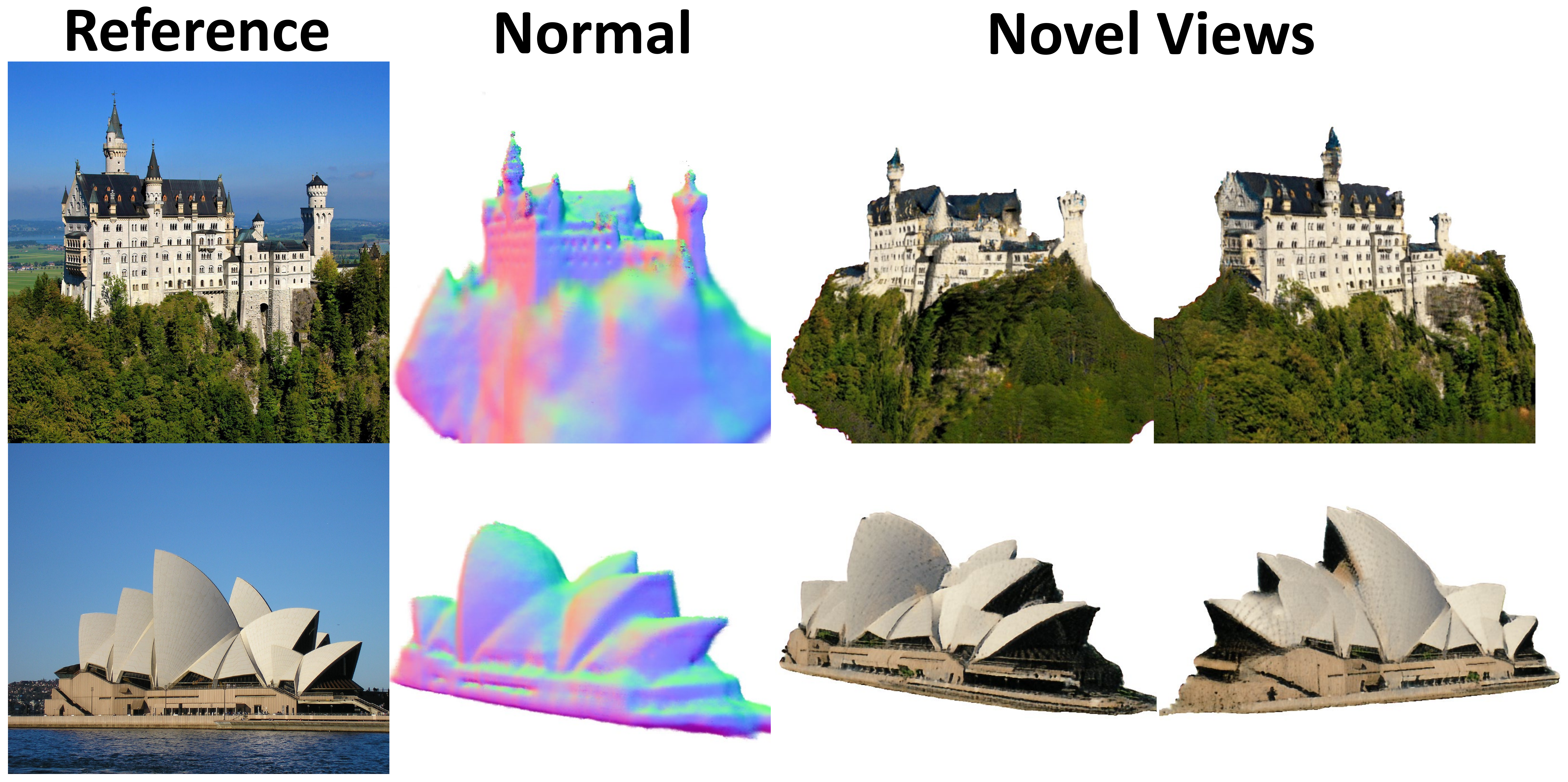}
    \caption{\textit{Make-It-3D} enables high-fidelity 3D creation on real complex scenes. }
    \label{fig:real}
    \vspace{-3mm}
\end{figure}

\begin{figure}
    \centering
    \includegraphics[width=\linewidth]{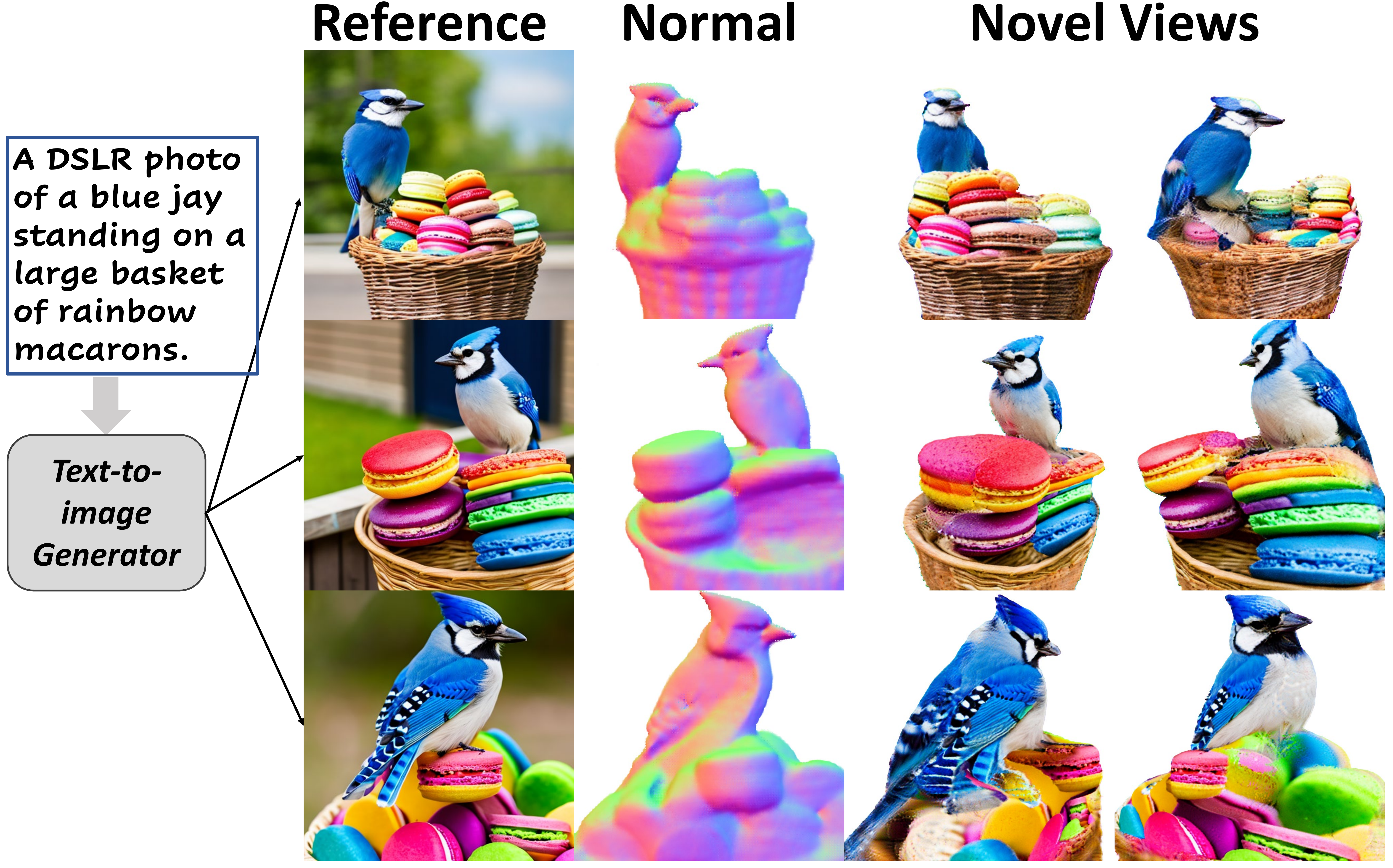}
    \caption{\textit{Make-It-3D} generates diverse and visually stunning 3D models given a text description.}
    \label{fig:app-1}
\end{figure}

\begin{figure}
    \centering
    \includegraphics[width=\linewidth]{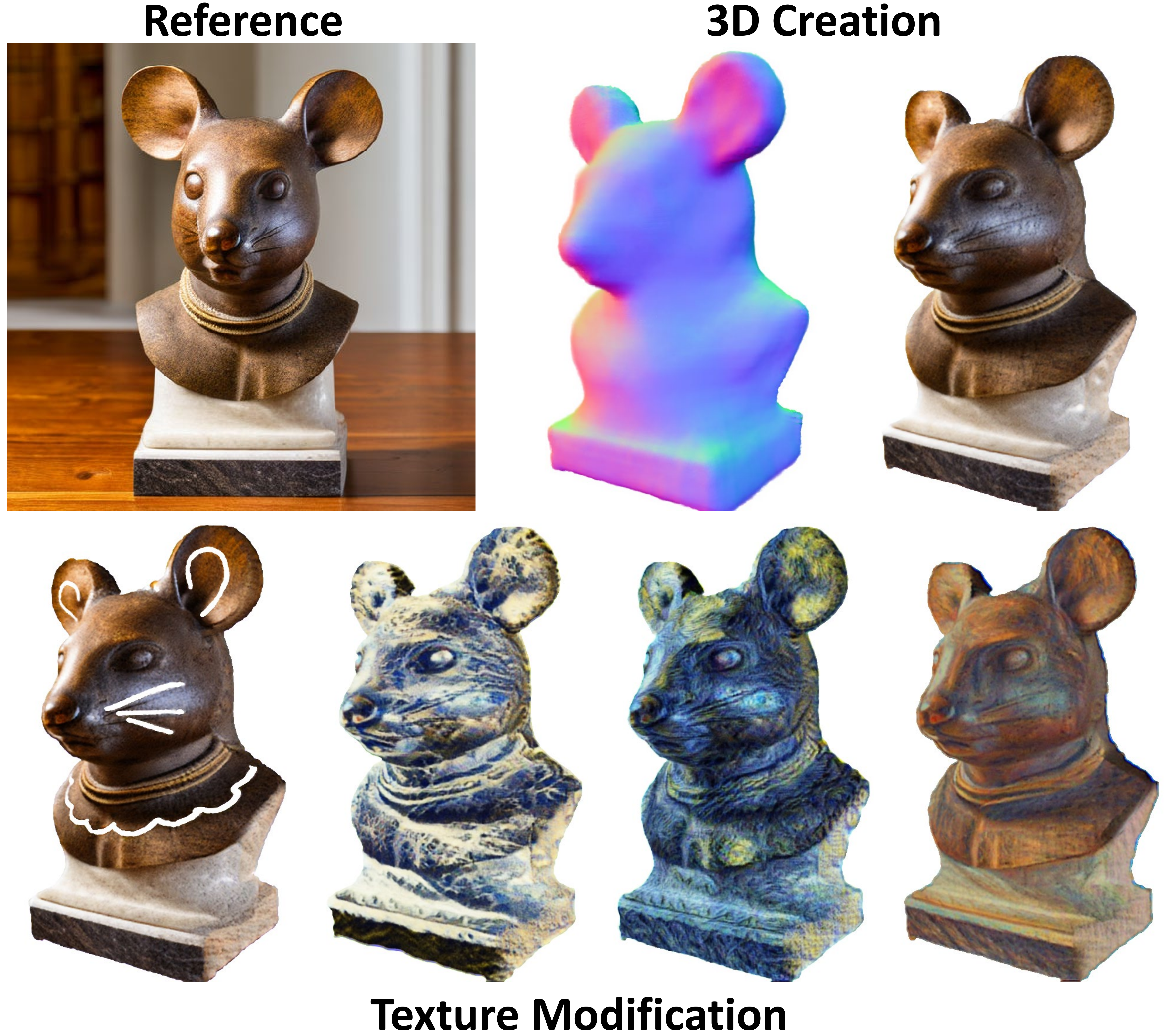}
    \caption{\textit{Make-It-3D} achieves 3D-aware texture modification such as tattoo drawing and stylization.}
    \label{fig:app-2}
    \vspace{-2mm}
\end{figure}

\section{Applications}
\noindent\textbf{Real scene modeling.} As shown in Figure~\ref{fig:real}, \emph{Make-It-3D} can successfully convert  a single photo of a complex scene to a 3D model, such as buildings and landscapes. This empowers users to model a scene with ease, which could be difficult for some traditional 3D modeling techniques.

\noindent\textbf{High-quality  text-to-3D generation with diversity.} Prior arts~\cite{poole2022dreamfusion,lin2022magic3d} often produce models with limited diversity and excessively smooth textures. To perform high-quality text-to-3D creation, we first convert the text prompt to a reference image using 2D diffusion, and proceed with our image-based 3D creation method.  As shown in Figure~\ref{fig:app-1}, \emph{Make-It-3D} is capable of producing diverse examples from a text prompt that exhibit stunning quality.   

\noindent\textbf{3D-aware texture modification.} \emph{Make-It-3D} enables view-consistent texture editing by manipulating  the reference image in the refine stage while freezing the geometry. Figure~\ref{fig:app-2} shows that we can add a tattoo and apply stylization to the generated 3D model.

\section{Conclusions}
We introduce  \textit{Make-It-3D},   a novel two-stage method for creating high-fidelity 3D content from one single image. 
Leveraging diffusion prior as 3D-aware supervision, the generated 3D models exhibit faithful geometry and realistic textures with the diffusion CLIP loss and textured point cloud enhancement. \textit{Make-It-3D}  
is applicable to general objects, empowering versatile fascinating applications. 
We believe our method takes a big step in extending the success of 2D content creation to 3D, providing users with a fresh 3D creation experience.


{\small
\bibliographystyle{ieee_fullname}
\bibliography{egbib}
}

\clearpage
\section*{Appendix}
\appendix
\section{Broad Impact}
We have presented \emph{Make-It-3D}, a novel approach to create novel views from a single image of general genre. Make-It-3D first hallucinates the 3D geometry by the usage of depth prior at the frontal view and the geometry prior of a pretrained diffusion model to ensure plausibility at novel views. Motivated by the fact that human eyes are more sensitive to texture over geometry, we thus reuse the coarse 3D geometry estimated from the implicit representation as well as the texture from the reference image, and specifically ``inpaints'' the texture of explicit 3D representation at occluded regions, ultimately producing compelling novel view renderings with highly-detailed texture. 

Our primary aim is to advance the research of generative modeling from 2D to 3D. Without relying on 3D training data that is hardly accessible in scale, this work tackles the 3D synthesis problem by lifting 2D generated images to 3D. This way essentially builds on the assumption that a diffusion model not only generates 2D observations but also implicitly contains rich 3D understanding of the scene. Thus, using our technique, one can generate a 3D scene that can be immersively viewed by merely using a 2D diffusion model. Compared to DreamFusion and Magic3D, our work produces more diverse 3D synthesis results with significantly improved realism. On top of creatively generated images, this work also performs well on real images with complicated structures.

We hope this work opens the door towards high-quality 3D synthesis and inspires more following works along this way. While we have demonstrated the ability to synthesize novel views in 360 degree, it is still non-trivial to produce holistically plausible 3D objects when viewed from large viewpoints.  Moreover, while this work aims for 3D synthesis from a single image, the same pipeline is applicable to the few-shot scenario where a few multi-view images can be obtained.  In addition, it would be fruitful to generalize the proposed technique to augment the quality of 4D synthesis. We will release the code to facilitate the research in this emerging area. 

\section{Additional Implementation Details}
\subsection{Coarse stage}
\noindent\textbf{Scene representation and rendering.}
We use the explicit-implicit representation from Instant-NGP~\cite{muller2022instant} to implement the NeRF representation in the coarse optimization stage, where we choose 16-level hash encoding of size $2^{19}$ and dimension 32, with a 3-layer MLP with 64 hidden units to decode the density and color for each spatial location.
During volumetric rendering, we sample 96 points for each ray, including 64 points for uniform sampling and 32 for importance sampling. We initialize the density field as a Gaussian sphere, which leads to faster convergence and more stable training. Specifically, we initialize the density as
$\sigma_{\textnormal{init}} = d * \exp(-||x||^{2} / (2\mu^{2}))$,
where we set density bias $d=5$ and $\mu=0.2$; $x$ denotes the distance between the ray point and the scene center. 

\noindent\textbf{Camera setting.} Following the camera sampling method used in~\cite{poole2022dreamfusion}, we randomly sample camera distance from 0.8 to 1.2, and the field-of-view (FOV) from 40 to 80 degrees. We find that randomly sampling FOV is instrumental to mitigate the artifacts that arise in large rendering view angles.
%

\begin{figure*}[htb!]
    \centering
    \includegraphics[width=\linewidth]{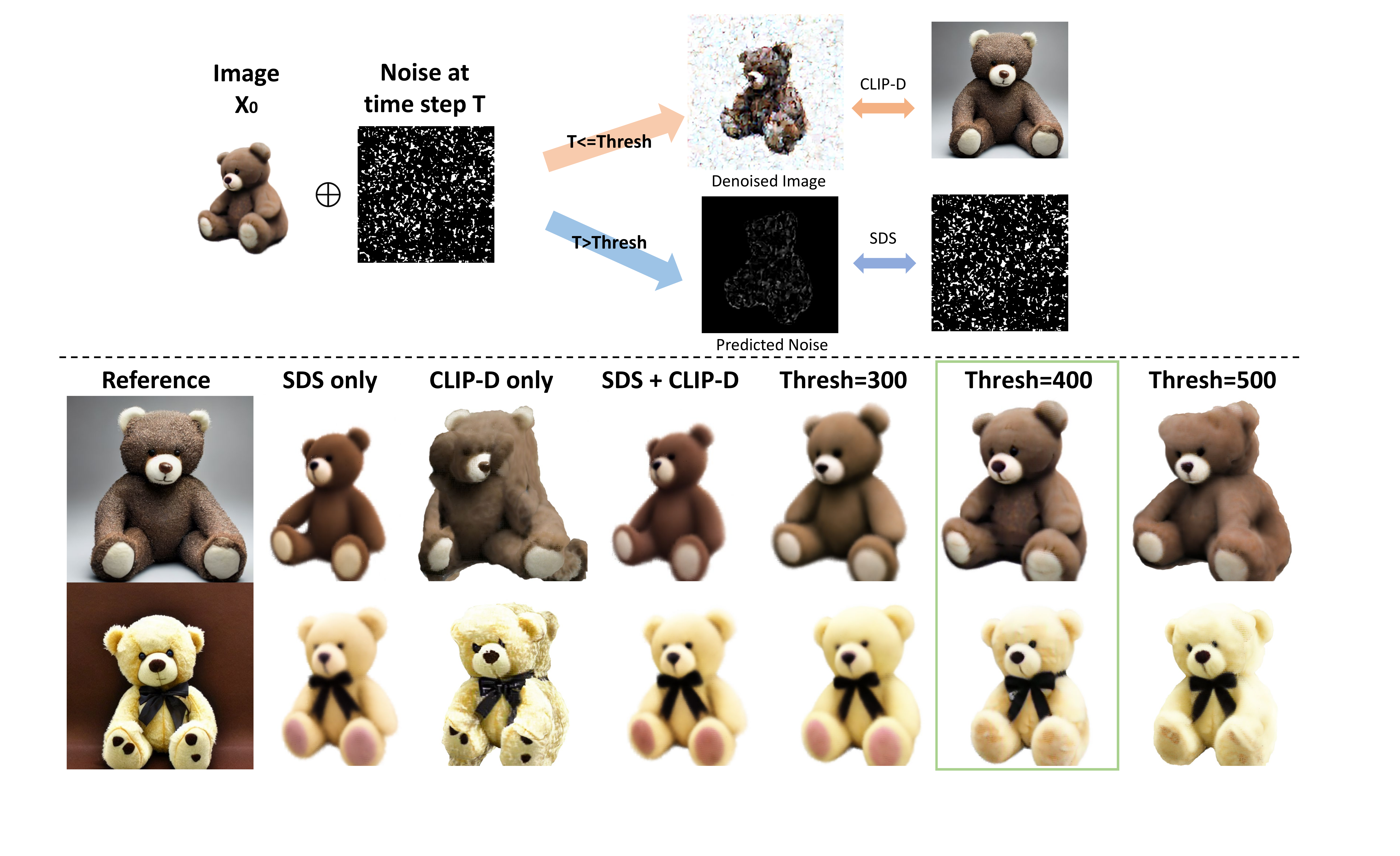}
    \caption{Analysis of SDS and CLIP-D loss.}
    \label{fig:clip-d}
\end{figure*}

\noindent\textbf{Augmentation and Regularization.}
To encourage the network to focus more on the foreground and avoid adversarial samples that hack the pretrained diffusion model, we train NeRF with a random background augmentation. Specifically, during training, we randomly jitters the background color of both the reference alpha image and NeRF rendering. During inference, we render the scene with a white background. 
Furthermore, following~\cite{poole2022dreamfusion}, we use three types of geometric regularization including sparsity, opacity and smoothness.

\subsection{Refine stage}

\noindent\textbf{Point cloud rasterization.}
Following~\cite{aliev2020neural}, we rasterize neural points $V$ to multi-scale feature maps $\mathcal{S}(i, V),~i\in[0, K)$, $K=3$. We use a differentiable point rasterizer implemented by PyTorch3D~\cite{ravi2020accelerating} to assign every pixel a neural descriptor and a binary scalar that indicates a non-empty pixel. We consider the binary mask as a point-based occupancy mask. 

\noindent\textbf{Background regularization.}
To handle pixels without corresponding point cloud projection, we assign a learnable descriptor as the background.  During texture enhancement optimization, we additionally add a regularization to encourage the scene to be rendered with a white background according to the binary occupancy mask mentioned above.

\noindent\textbf{Deferred neural rendering.}
For deferred rendering of the point clouds, we use a 2D U-Net architecture with gated convolutions~\cite{yu2019free}. It contains 3 down- and up-sampling layers to integrate multi-scale feature maps and output the final RGB image.








\begin{table}[t]
\centering\footnotesize
\begin{tabular}{l|ccc}
\toprule
 & LPIPS$\downarrow$  & Contextual$\downarrow$ & CLIP$\uparrow$ \\ \midrule
SDS & 0.3045 & 2.29& 86.04$\%$ \\
CLIP-D & \textbf{0.1260} & 2.43& 80.27$\%$\\
\midrule
SDS+CLIP-D& 0.2772 &  2.32& 84.01$\%$  \\
Thresh=300& 0.1757 & 2.19& 87.40$\%$ \\
Thresh=400& 0.1427 & \textbf{1.74}& \textbf{87.50$\%$} \\
Thresh=500& 0.1696 & 2.23& 86.09$\%$ \\

\bottomrule
\end{tabular}
\caption{Ablation study on  SDS and CLIP-D loss on the test benchmark. We compute LPIPS under the reference view, and the other two metrics under novel views. ``Thresh" denotes the boundary of time steps using SDS or CLIP-D in the denoising process.}
\label{tab:clip-d}
\end{table}

\section{Additional Ablation Study and Analysis}

\subsection{Analysis of SDS and CLIP-D loss}

As mentioned in Sec {\color{red}3.1}, in the coarse stage, we use the diffusion prior by applying score distillation sampling (SDS) scheme on novel view renderings. It can successfully encourage the generated scene to match the conditioned text prompt.
However, as an image-based 3D content creation model, we need to prioritize the faithfulness between created 3D and the reference image. Although we add pixel-wise constrain under the reference view for optimization, SDS provides a strong geometric prior and enforces the optimized scene to be a plausible result according to the text condition. Constraints under a single view can be limited. Thus the created results may not be rigorously aligned with the reference image (See Figure~\ref{fig:clip-d}).

Therefore, we need to relax the strong geometric guidance provided by SDS and add more image-level constraints under multi-views.
We achieve this goal by simultaneously maximizing the image-level similarity between the reference image and the novel view renderings denoised by the diffusion model, named as a diffusion CLIP loss $\mathcal{L}_\textnormal{CLIP-D}$.
Compared with introducing this constraint directly on novel view renderings, the CLIP-D encourages the pretrained diffusion model to provide better guidance to generate more faithful 3D content with the reference image.

In view of this, we conduct several experiments to study the effect of SDS and CLIP-D loss during optimization, which is shown in Figure~\ref{fig:clip-d}. Results show that using only SDS generates high-quality and plausible geometry, but the optimized 3D does not align with the image. On the contrary, using only CLIP-D preserves the appearance of the reference image, but fails to generate good geometry. A simple solution is to combine the two losses, but this does not fully address the non-alignment issue. 
To achieve a balance between geometric quality and appearance alignment, we introduce an optimization strategy by setting a threshold of sampling steps. Specifically, we optimize CLIP-D loss at small timesteps and optimize SDS at large steps. 
We conduct several qualitative and qualitative studies on different threshold settings, which are shown in Figure~\ref{fig:clip-d} and Table~\ref{tab:clip-d}. 
During training, we randomly sample noise step $T$ from 200 to 600,
and we find that $T=400$ could balance the geometric quality and the appearance alignment.

\begin{figure}
    \centering
    \includegraphics[width=\linewidth]{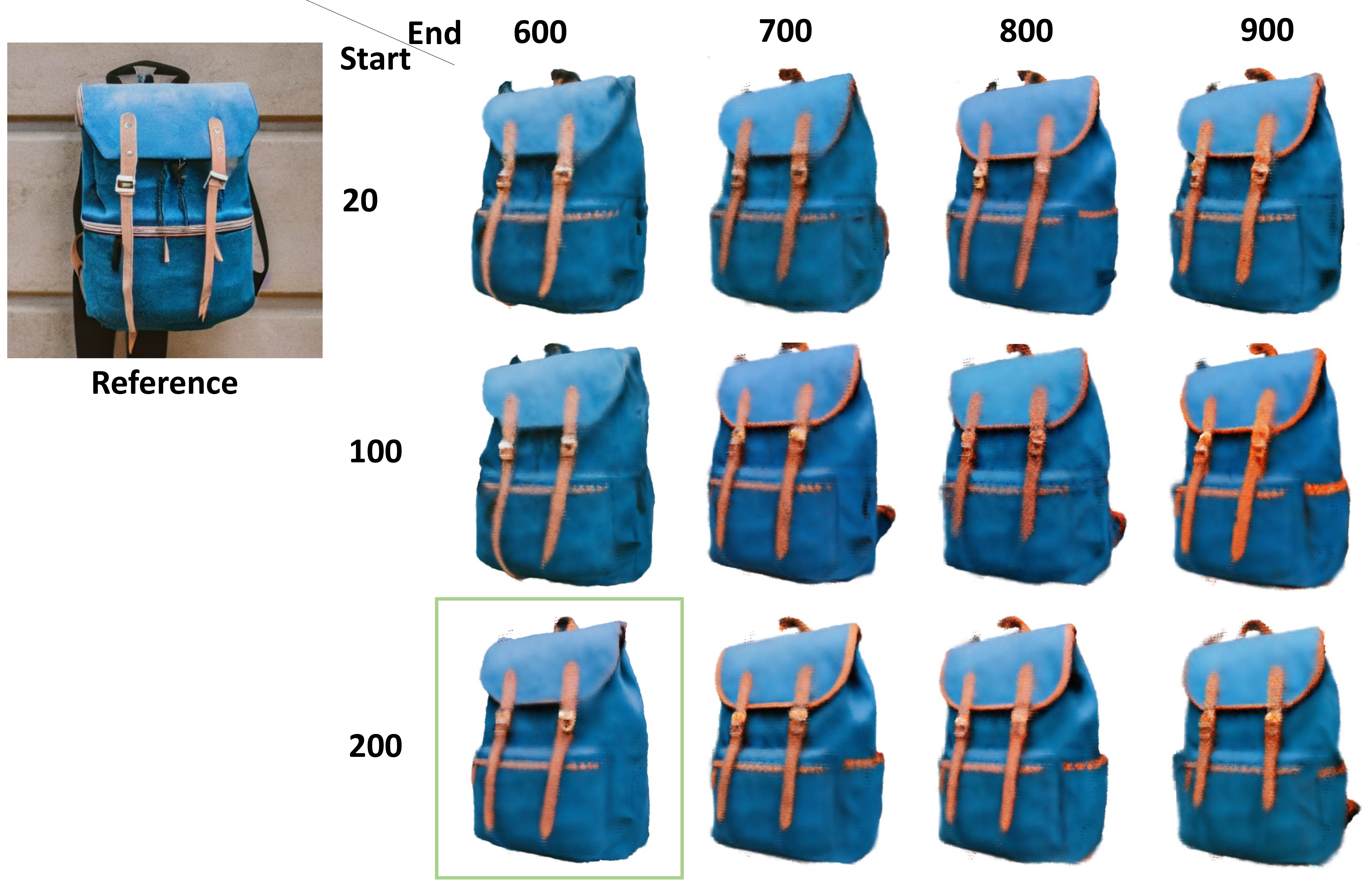}
    \caption{Analysis of the time step range  in SDS process. We visualize  novel view results in the coarse stage that are trained with different time step ranges (from start to end).}
    \label{fig:step}
    \vspace{-3mm}
\end{figure}

\begin{figure}
    \centering
    \includegraphics[width=\linewidth]{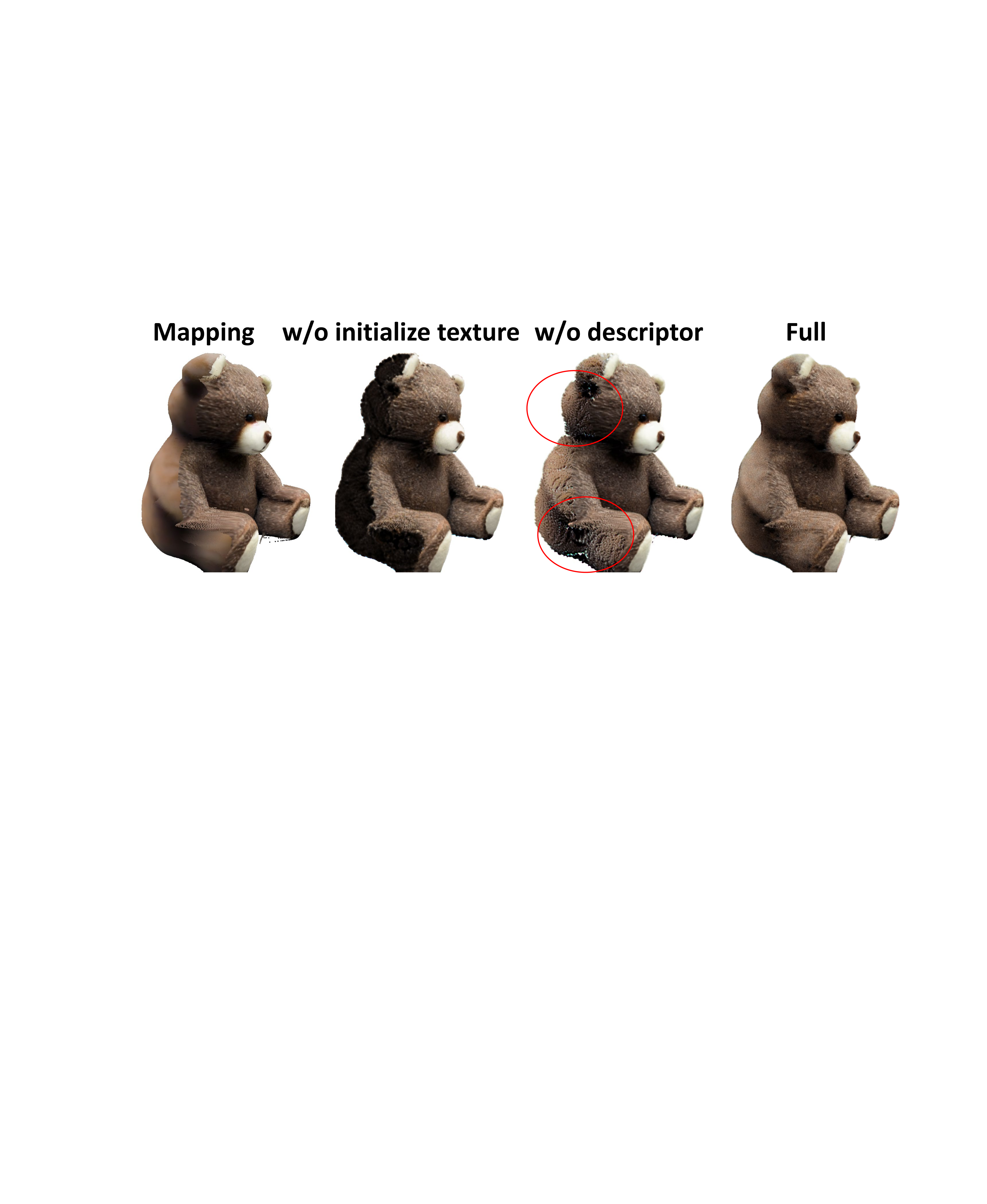}
    \caption{Analysis of texture initialization and point descriptors.}
    \label{fig:ab-2}
    \vspace{-4mm}
\end{figure}

\subsection{Analysis of various sampling time step ranges}

We also investigate the effect of various sampling time steps in SDS process. The experimental results are shown in Figure~\ref{fig:step}. We conduct several experiments using different sampling ranges. We observe that adding noise at large time steps can improve the geometry quality but reduce the alignment and potentially saturate textures. And the diffusion prior does not provide adequate supervision at small time steps. In our method, we exclude small and large time steps and instead randomly sample time step $T$ from 200 to 600.

\subsection{Analysis of texture initialization and point descriptors}
We conduct ablation studies on texture enhancement process. We explore the importance of the initialized unseen texture from NeRF and point descriptor. The qualitative results are shown in Figure~\ref{fig:ab-2}. We can see that texture initialization is crucial for global texture enhancement. And only optimizing point color without descriptor outputs artifacts and cannot produce reasonable results. 

\begin{figure}
    \centering
    \includegraphics[width=0.9\linewidth]{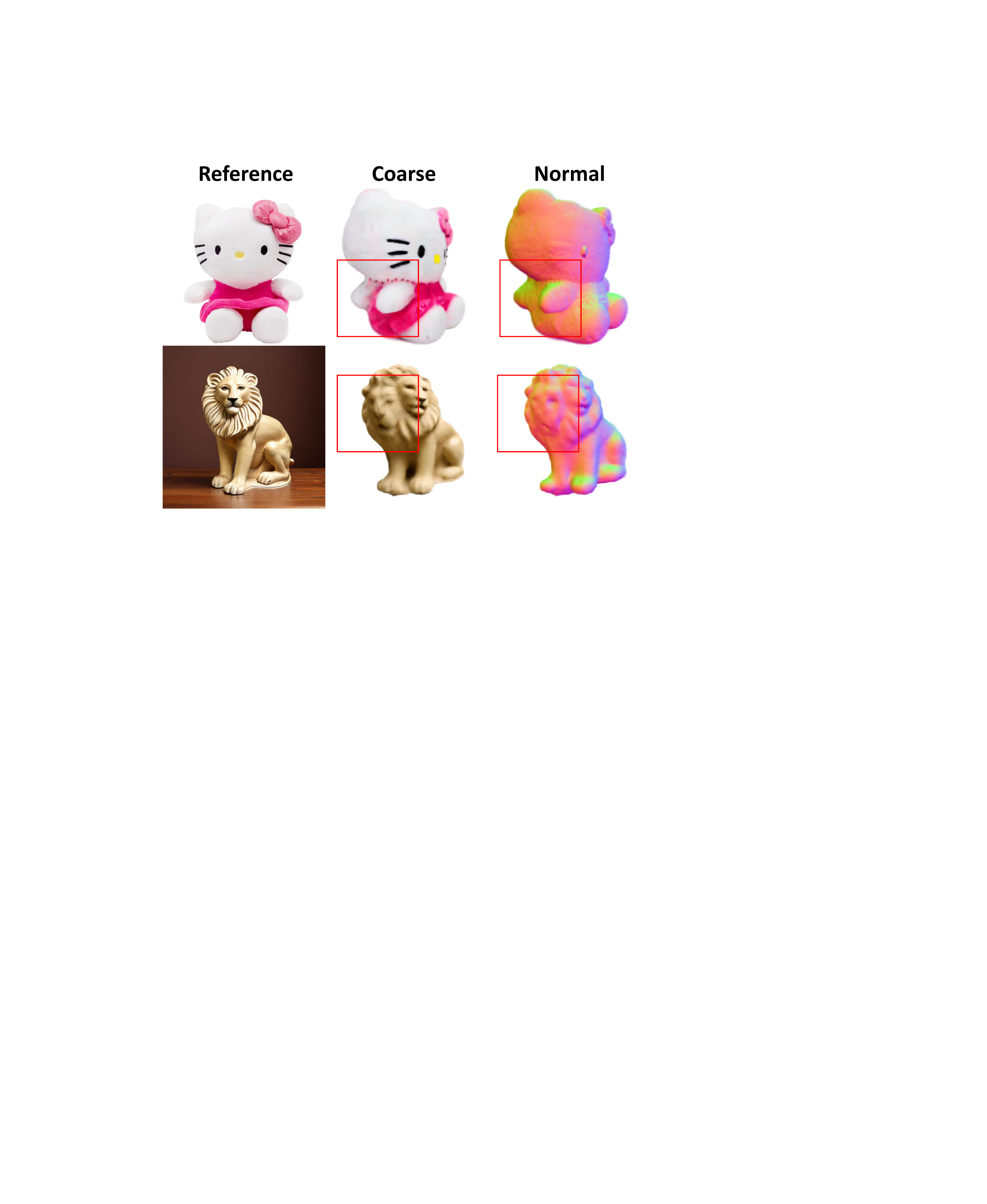}
    \caption{Failure cases due to the geometry ambiguity.}
    \label{fig:limit}
    \vspace{-3mm}
\end{figure}

\begin{figure*}[htb!]
    \centering
    \includegraphics[width=\linewidth]{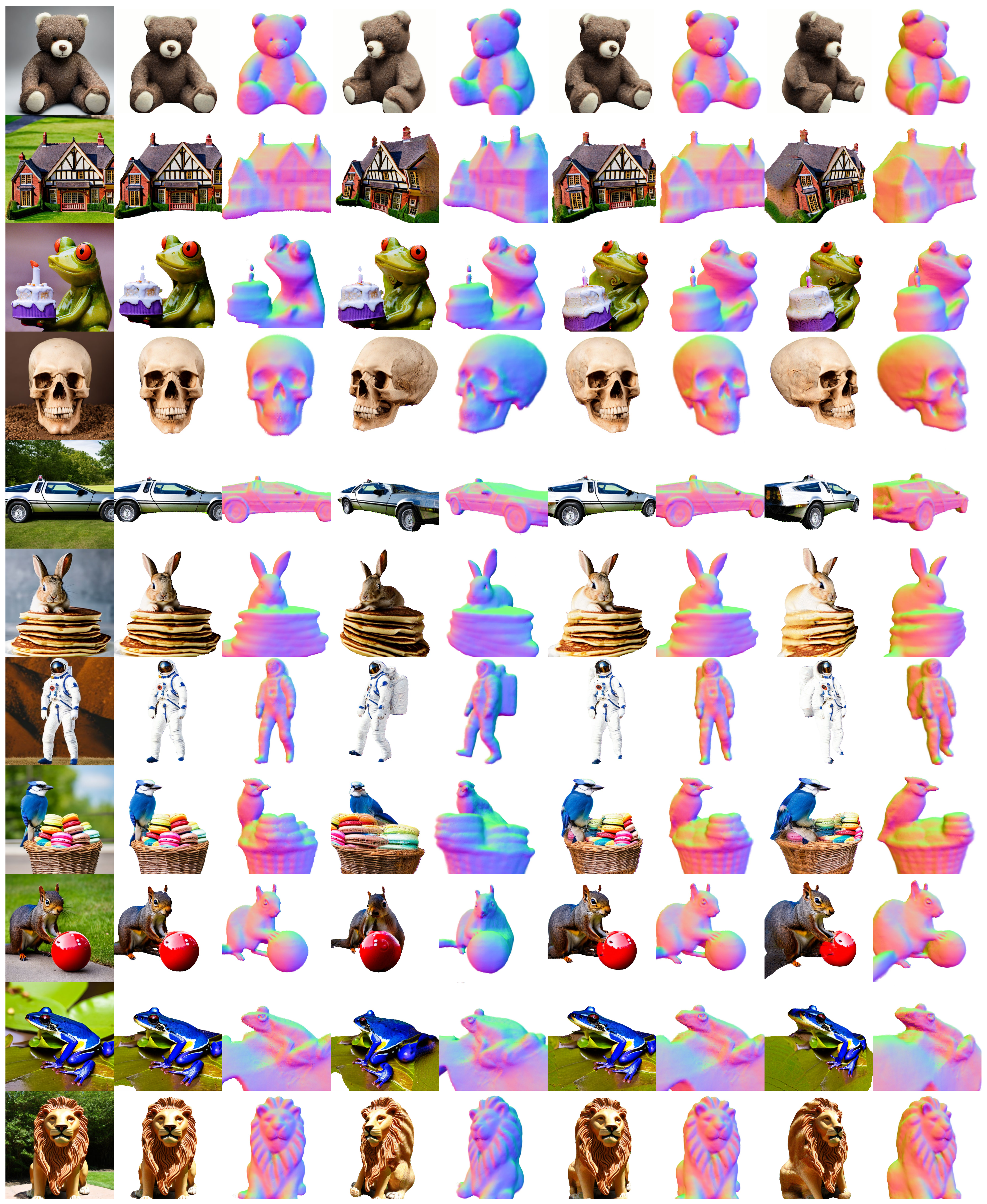}
    \caption{Additional results by \textit{Make-It-3D}. The first column is the reference image. We show high-fidelity results including normal maps under the reference view and novel views.}
    \label{fig:more}
\end{figure*}

\begin{figure*}[htb!]
    \centering
    \includegraphics[width=\linewidth]{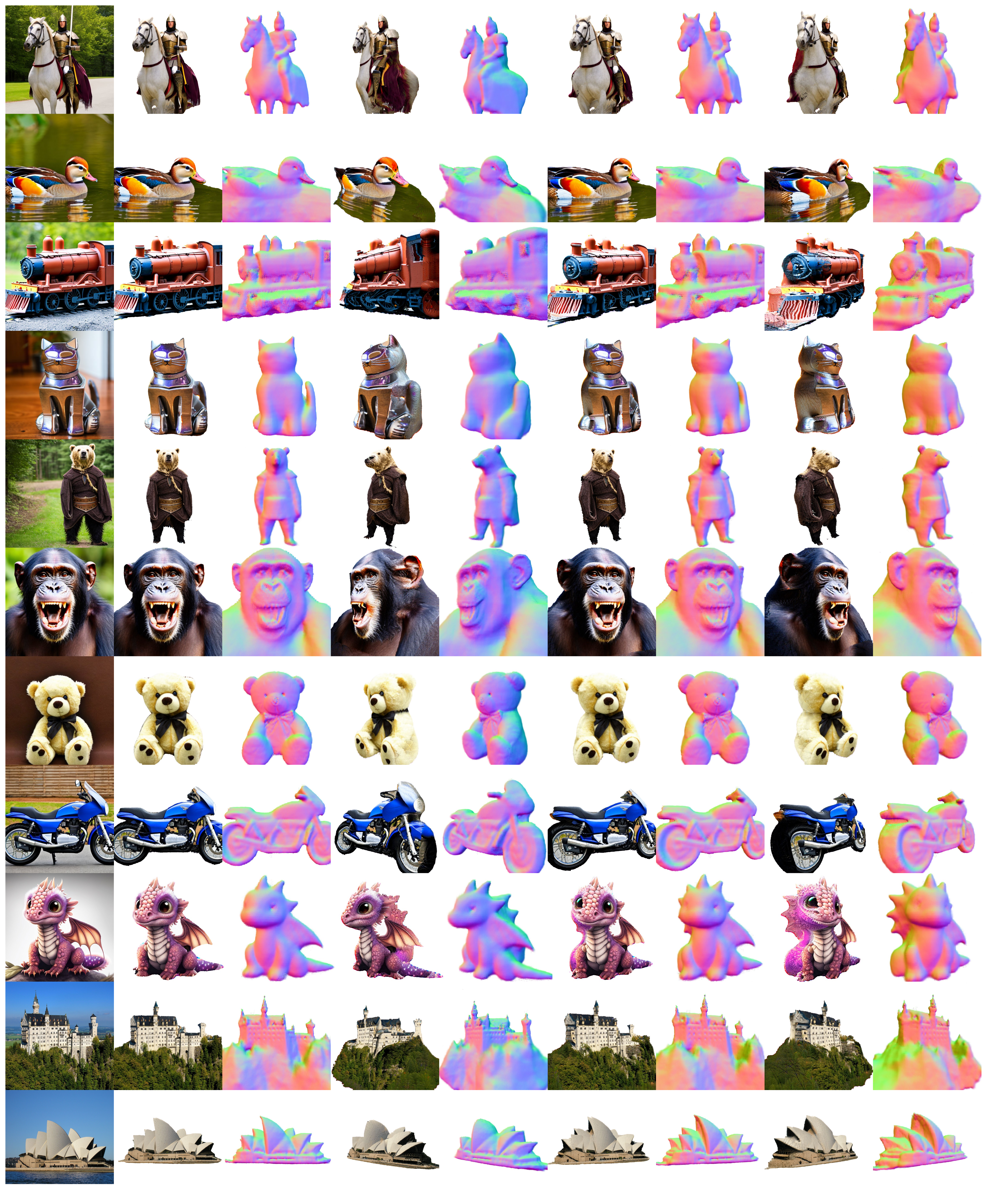}
    \caption{Additional results by \textit{Make-It-3D}. The first column is the reference image. }
    \label{fig:more-2}
\end{figure*}

\begin{figure*}[t]
    \centering
    \includegraphics[width=\linewidth]{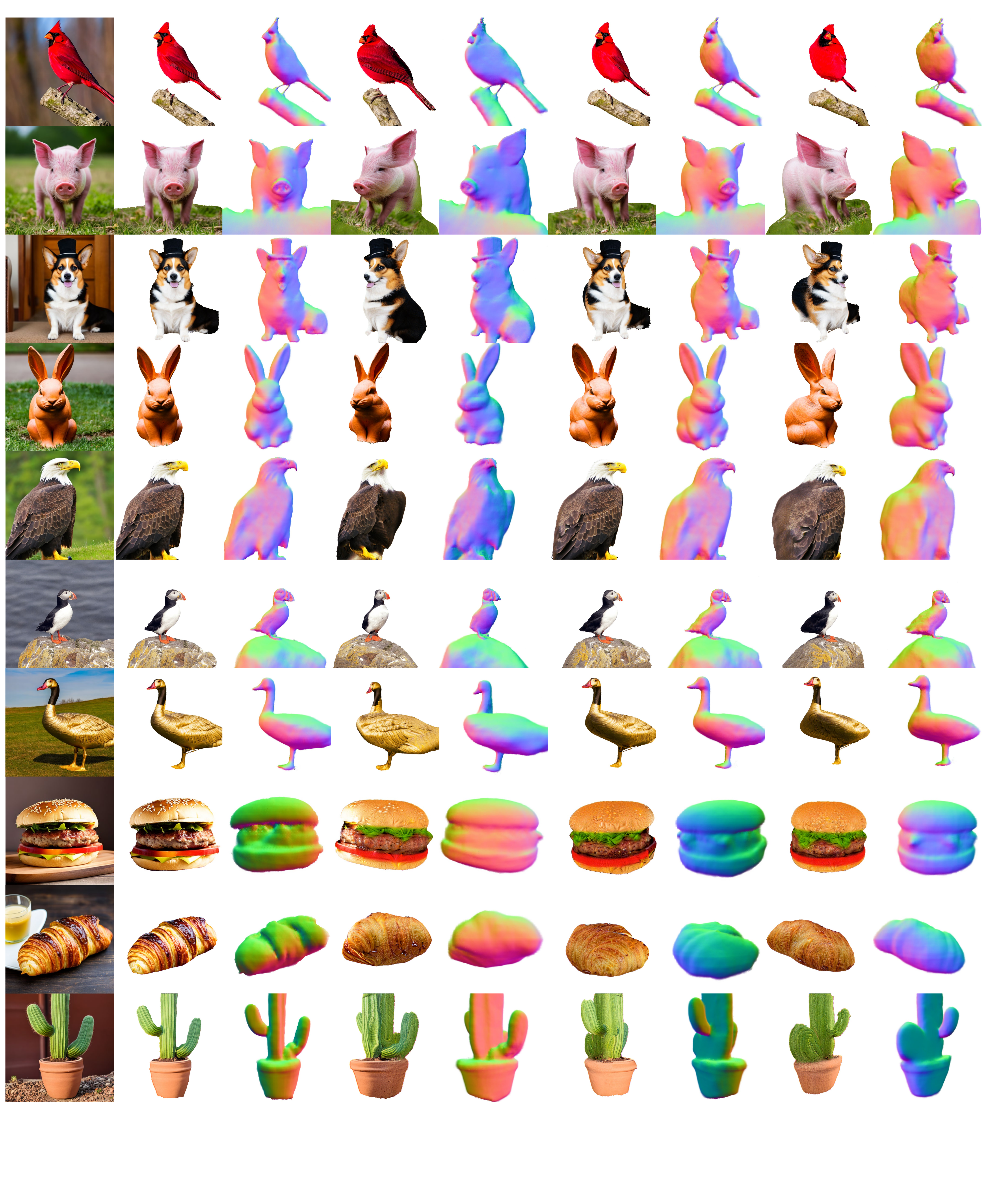}
    \caption{Additional results by \textit{Make-It-3D}. The first column is the reference image. }
    \label{fig:more-3}
\end{figure*}

\section{Additional Results}

In this section, we provide additional results of creating 3D models from different reference images using our method.
The results are shown
on Figure~\ref{fig:more}, Figure~\ref{fig:more-2}, and Figure~\ref{fig:more-3}. Results show that our method has a strong ability on creating high-fidelity 3D content including high-quality geometries and textures using a single reference image.

\section{Limitations}
Our method suffers from some geometry ambiguity, such as Janus problem or over-flat geometry~\cite{poole2022dreamfusion}. A depth prior can reduce this issue. However, since we only add depth constrain at a single view, the geometry ambiguity may still exist under other views. We show some failure cases in Figure~\ref{fig:limit}.


\end{document}